\documentclass[letterpaper, 10 pt, conference]{ieeeconf}  

\IEEEoverridecommandlockouts                              

\usepackage{graphics} 
\usepackage{epsfig} 
\usepackage{mathptmx} 
\usepackage{times} 
\usepackage{amsmath} 
\usepackage{amssymb}  
\usepackage{subfigure}
\usepackage{multirow}
\usepackage{subcaption}
\usepackage{float}
\usepackage{ctable}
\usepackage{cuted}
\usepackage{colortbl}
\usepackage{algorithm}
\usepackage{algpseudocode}
\usepackage{titlesec}
\usepackage{soul}

\usepackage[numbers,sort&compress]{natbib}
\usepackage{eucal} 
\usepackage[colorlinks,linkcolor=blue]{hyperref}
\title{\Large \bf
Human-Agent Joint Learning for Efficient Robot Manipulation Skill Acquisition
}
\DeclareMathAlphabet{\mathcal}{OMS}{cmsy}{m}{n}

\author{\begin{tabular}{c}
    Shengcheng Luo\textsuperscript{1*}, Quanquan Peng\textsuperscript{1*}, Jun Lv\textsuperscript{1*}, Kaiwen Hong\textsuperscript{2}, \\
    \newline
  Katherine Rose Driggs-Campbell\textsuperscript{2}, Cewu Lu\textsuperscript{1}, Yong-Lu Li\textsuperscript{1} \\
  \end{tabular}
\thanks{* denotes equal contribution }
\thanks{$^{1}$Shanghai Jiao Tong University, $^{2}$University of Illinois Urbana-Champaign.}
}

\begin{document}

\maketitle
\thispagestyle{empty}
\pagestyle{empty}

\begin{strip}
\vspace{-15mm}
\centering
\includegraphics[width=1.0\textwidth]{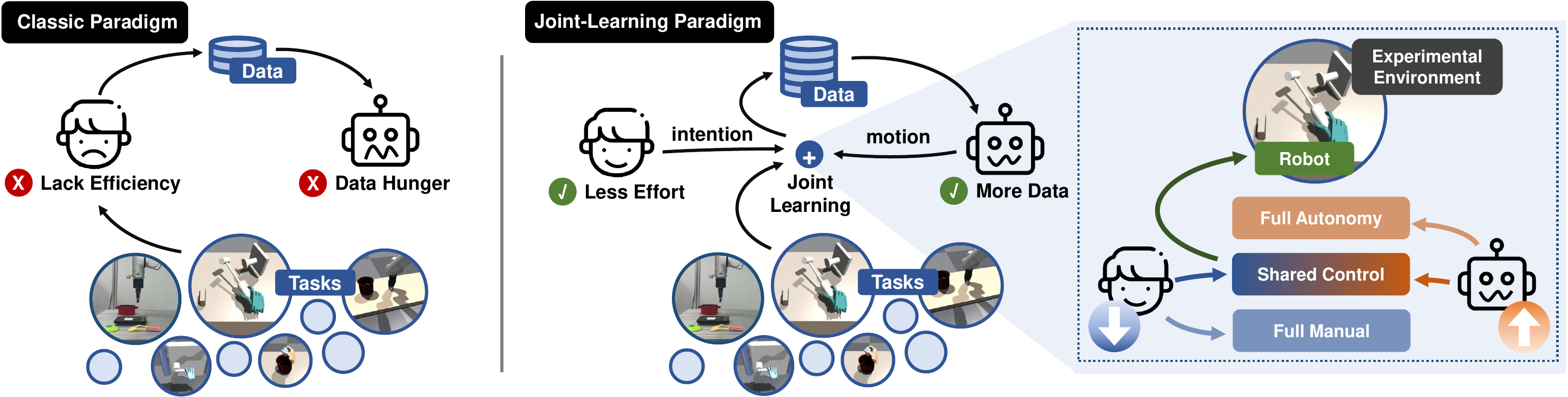}
\captionof{figure}{\textbf{Human-agent joint learning overview.} Traditional frameworks typically separate human and agent training, requiring operators first to learn the task environment before data collection. This often leads to inefficiencies due to delayed and insufficient data gathering. In our framework, we integrate human and agent training from the start in a joint learning model. This enables simultaneous development and adapts the agents to human operation more effectively, enhancing overall efficiency and promoting better collaboration between humans and machines allowing for human effortless adaptation data collection.}\label{fig:human_centric}
\vspace{-10px}
\end{strip}

\begin{abstract}
Employing a teleoperation system for gathering demonstrations offers the potential for more efficient learning of robot manipulation. However, teleoperating a robot arm equipped with a dexterous hand or gripper, via a teleoperation system presents inherent challenges due to the task's high dimensionality, complexity of motion, and differences between physiological structures.
In this study, we introduce a novel system for joint learning between human operators and robots, that enables human operators to share control of a robot end-effector with a learned assistive agent, simplifies the data collection process, and facilitates simultaneous human demonstration collection and robot manipulation training. As data accumulates, the assistive agent gradually learns. Consequently, less human effort and attention are required, enhancing the efficiency of the data collection process. It also allows the human operator to adjust the control ratio to achieve a trade-off between manual and automated control.
We conducted experiments in both simulated environments and physical real-world settings. Through user studies and quantitative evaluations, it is evident that the proposed system could enhance data collection efficiency and reduce the need for human adaptation while ensuring the collected data is of sufficient quality for downstream tasks. \textit{For more details, please refer to our webpage \href{https://norweig1an.github.io/HAJL.github.io/}{https://norweig1an.github.io/HAJL.github.io/}}.
\end{abstract}

\section{INTRODUCTION}
In recent years, significant progress has been made on learning robot manipulation policies from demonstrations. Previous studies have utilized teleoperation systems~\cite{arunachalam2023holo,gharaybeh2019telerobotic,liu2017glove, liu2019high,handa2020dexpilot,qin2023anyteleop} to collect human demonstrations, and learning-based policies~\cite{mandlekar2021matters,chi2023diffusion,florence2022implicit} have been formulated using the gathered data. Despite the notable advancements, several challenges still need to be addressed.
For example, in vision-based teleoperation systems, even with state-of-the-art 3D hand pose estimation algorithms~\cite{lv2021handtailor, rong2021frankmocap, schmidt2014dart,weichert2013analysis}, errors persist that significantly degrade the teleoperation. Additionally, discrepancies between the structures of human hands and robot end-effectors, along with the lack of haptic feedback during contact-rich manipulation, also pose challenges. As a result, current teleoperation systems demand substantial human effort, in addition, collecting high-quality datasets remains a challenging and labor-intensive task in many scenarios. 


    Naturally, a question was raised: \textit{in data-collection, how to make human effort less while improving the data quality?}
    Here, we aim to address this question and argue that human-agent joint learning can help.
    That said, an effective and efficient teleoperation system should be designed to preferentially capture the operator's intentions for directing a robot end effector and pose the \textit{main frame}, while concurrently enabling an autonomous agent to help us ensure motion stability and \textit{interpolate} the details.
    To this end, we propose a framework that achieves shared control between the human and a learned assistive agent during data collection. 
    As shown in Fig.~\ref{fig:human_centric}, our \textit{human-agent joint learning} framework seeks to integrate the data collection and policy learning 
    to enhance the efficiency of the whole process, reducing human effort, and improving the data quality.
    
    Given our human-agent joint learning approach, 
    we allow the data acquisition agent to grow and learn along with the human operator.
Inspired by shared autonomy~\cite{javdani2015shared,reddy2018shared,schaff2020residual}, 
    we introduce a novel teleoperation system that enables collaboration between humans and learning-based agents to control a robot jointly during the data collection and learning process. 
    In particular, our proposed system provides the flexibility to adjust a ``control ratio'' between the human operator and a learning-based agent. 
    A lower control ratio, in the beginning, emphasizes the human's role in teaching the agent finer-grained knowledge under the structure of human intention and principal actions.
    As the agent's learning improves, a higher ratio indicates greater autonomy from the learned agent to replace the human effort to ``inpaint'' the whole process given only human intention and principal actions. 
    

With the proposed system, the human effort will be reduced due to the shared control during data collection. Additionally, the agent learning process is integrated with the data collection, improving the efficiency of the whole process. In addition, the quality of the collected data is also improved, benefiting different kinds of downstream tasks.
    We conducted experiments in six different simulation environments using two types of end-effectors: a dexterous hand and a gripper. Additionally, we performed experiments on three real-world tasks to validate our findings. Evaluation results indicate that our proposed system significantly enhances data collection efficiency, increasing the collection success rate by 30\% and nearly doubling the collection speed. 
    Additionally, data collected in shared autonomy mode is as effective for downstream tasks and models as data collected directly from the teleoperation system, demonstrating comparable validity. 
    Our main contributions are summarized as follows:
\begin{itemize}
    \item We study how to reduce human adaptation while keeping data quality in teleoperation data collection and propose a human-agent joint learning paradigm. 
    \item We build a system 
    that fosters concurrent development between the human operator and assistive agent, which not only streamlines the learning process but also expedites the robot's ability to perform robot manipulations autonomously.
    \item Conducting both simulation and real-world experiments to demonstrate the efficiency and effectiveness of our proposed system. Our system achieved significant performance improvements, including a \textbf{30\%} increase in data collection success rate and \textbf{double} the collection speed. 
\end{itemize}

\section{RELATED WORKS}


    \textbf{Teleoperation for Data Collection.} Data has always been a crucial foundation, and robots are no exception. Teleoperation serves as a significant source for collecting robot data~\cite{mandlekar2021matters, brohan2022rt,ebert2021bridge,fang2023rh20t,kofman2005teleoperation,mandlekar2018roboturk,fang2024airexo}. 
    Some works achieve teleoperation through wearable devices~\cite{arunachalam2023holo,gharaybeh2019telerobotic,liu2017glove,liu2019high, lipton2017baxter}, and vision-based teleoperation systems offer a low-cost and easily developed alternative~\cite{handa2020dexpilot,qin2023anyteleop, antotsiou2019task, li2019vision}. 
    For instance, \cite{li2019vision} utilizes neural networks for markerless vision-based teleoperation of dexterous robotic hands from depth images. \cite{handa2020dexpilot} set up a vision-based teleoperation system to control the Allegro Hand, accomplishing various contact-rich manipulation tasks in the real world. Recently, \cite{qin2023anyteleop} introduced AnyTeleop, a unified teleoperation system designed to accommodate various arms, hands, realities, and camera setups within a singular framework. 
    In this paper, we introduce a joint learning paradigm to assist teleoperation by sharing control between the human operator and a learning-based agent, improving the efficiency of data collection using teleoperation.


    \textbf{Interactive robot learning.} Collecting fine-grained human demonstration data for robotic manipulation is an effective but labor-intensive and time-consuming way to enable robots to complete a wide range of tasks~\cite{liu2022robot, walke2023bridgedata}. Previous work uses shared autonomy to assist people with disability in performing tasks by arbitrating human inputs and robot actions~\cite{jeon2020shared}. Many of the shared autonomy algorithms aim to estimate human intents from a set of pre-defined goals~\cite{dragan2013policy, javdani2018shared, muelling2017autonomy, sadigh2016information}, using clothoid curves to parametrize the state and control~\cite{SkillRss} or by mapping low-dimension control input to high-dimension robot actions~\cite{jeon2020shared, losey2022learning}. 
    In this work, we introduce a system that integrates the agent's learning process with data collection, facilitating both data collection and robot learning.

\section{PROPOSED METHOD}

The primary contribution of this work is the development of a novel and highly efficient data collection method. To achieve this, the system is designed in two key stages: first, the proposed system allows human operators to control the robot via a teleoperation system to gather an initial but insufficient training dataset, which serves as the foundation for the second stage (Sec.~\ref{sec:vbts}). Second, using these data, we train a diffusion-model-based assistive agent (Sec.~\ref{sec:dp}) to establish shared control between the human operator and the agent, thereby improving the efficiency of the data collection process (Sec.~\ref{sec:process}). This approach mirrors the concept of ``bootstrapping''~\cite{chu2023bootstrappingroboticskilllearning}, where, as more data is accumulated, the system progressively reduces the effort required from human operators, facilitating further data collection and iterative system improvement. Additionally, once sufficient data has been gathered, the system offers the option to transition the shared control agent to full autonomy.

\subsection{Preliminary}
To get a learned agent in Sec.~\ref{sec:dp}, enabling human-agent joint learning, we follow the Denoising Diffusion Probabilistic Model (DDPM)~\cite{ho2020denoising} training paradigm. 
Here we first briefly introduce the DDPM algorithm.
The \textit{forward process} of the Diffusion Model can be regarded as adding Gaussian noise to the data $x^0$ according to a variance schedule $\beta_{1:K}$ by 
\begin{equation}
    x_k = \sqrt{\alpha_k}x_{k-1} + \sqrt{1 - \alpha_k} \epsilon,
    \label{fwd:diff}
\end{equation}
where  $\epsilon \sim \mathcal N (\mathbf{0, I}), \alpha_k=1-\beta_k$. DDPM models the output generation as a denoising process (Stochastic Langevin Dynamics). 
A line of works~\cite{janner2022planning, ajay2023conditional, chi2023diffusion, xu2023xskill} use diffusion model to generate the action for agents:
given $x^K$ sampled from Gaussian noise $\mathcal N (\mathbf{0, I})$, it utilizes a parameterized diffusion process to model how $x^K$ is denoised in order to get noise-free action $x^0$ by
\begin{equation}
    p_\theta(x^0)=\int p(x^K)\prod_{k=1}^K p_\theta(x^{k-1}|x^k) \mathrm d x^{1:K},
    \label{diff}
\end{equation}
where $p_\theta(x^{k-1}|x^k) = \mathcal N (\mu_\theta(x^k, k), \Sigma(x^k, k))$ is usually referred as \textit{reverse process}. \cite{luo2022understanding} shows that $p_\theta(x^{t-1}|x^k)$ becomes tractable when conditioned on $x_0$ and Eq.~\ref{diff} can be reformulated as minimizing the error in the noise prediction. \cite{ho2020denoising} simplify the training loss function as 
\begin{equation}
\mathcal{L}:=\mathbb{E}_{k,\boldsymbol{x}_0,\boldsymbol{\epsilon}\sim\mathcal{N}(\mathbf{0},\boldsymbol{I})}\left[\|\boldsymbol{\epsilon}-\boldsymbol{\epsilon}_\theta(\boldsymbol{x}_k(\boldsymbol{x}_0,\boldsymbol{\epsilon}),k)\|_2^2\right],
\label{loss}
\end{equation}
where step $k$ is sampled uniformly as $k \in [1, K]$, $\boldsymbol \epsilon_\theta$ is the noise prediction model. During the inference phase, we can generate $x_0$ by recursively sample $\boldsymbol{z} \sim \mathcal{N}(\mathbf{0},\boldsymbol{I})$:
\begin{equation}
    x_{k-1}=\mu_\theta(x_k, k) + \sigma_k \boldsymbol{z}.
\end{equation}

  Similar to \cite{chi2023diffusion, yoneda2023diffusha}, with the collected trajectory $\{(s_i,a_i)\}_{i=0}^T$, we aim to train an agent to imitate the trajectory, accomplishing a specific task $\mathcal{T}$. Therefore, we utilize DDPM to capture the conditional distribution of $p(a | s)$ and the training loss in Eq.~\ref{loss} shall be modified as 
\begin{equation}
\label{eqn:loss}
\mathcal{L}:=\mathbb{E}_{k,(s_i, a_i),\boldsymbol{\epsilon}\sim\mathcal{N}(\mathbf{0},\boldsymbol{I})}\left[\|\boldsymbol{\epsilon}-\boldsymbol{\epsilon}_\theta({a}_i + \boldsymbol{\epsilon}, s_i,k)\|_2^2\right].
\end{equation}

\vspace{-10px}

\begin{figure*}[t]
\centering
\resizebox{1\linewidth}{!}{\subfigure{
 \begin{minipage}[t]{0.98\linewidth}
    \includegraphics[width=\textwidth]{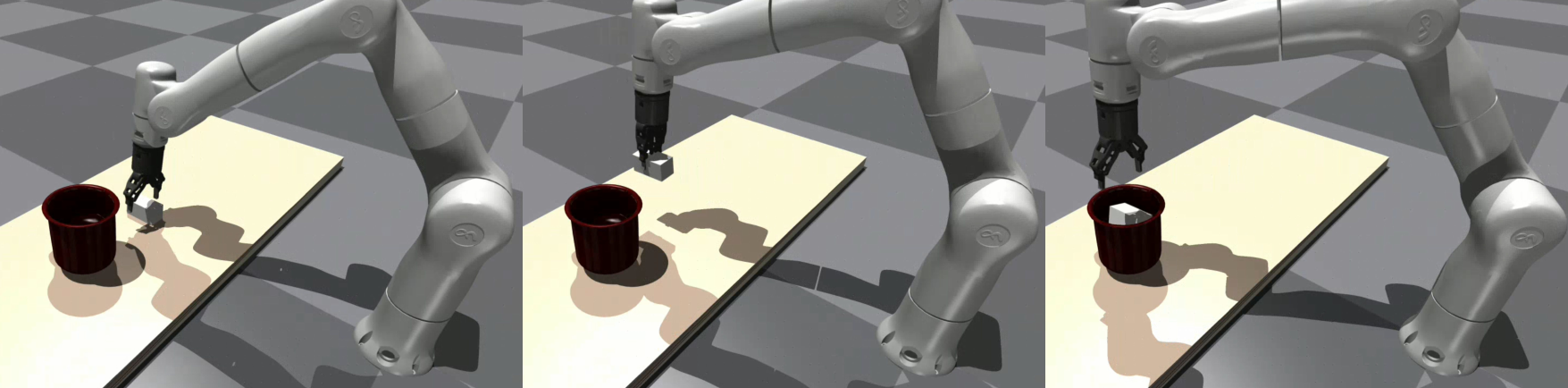}\label{fig:showp}
    \centering
    \end{minipage}}
\subfigure{
 \begin{minipage}[t]{0.98\linewidth}
    \includegraphics[width=\textwidth]{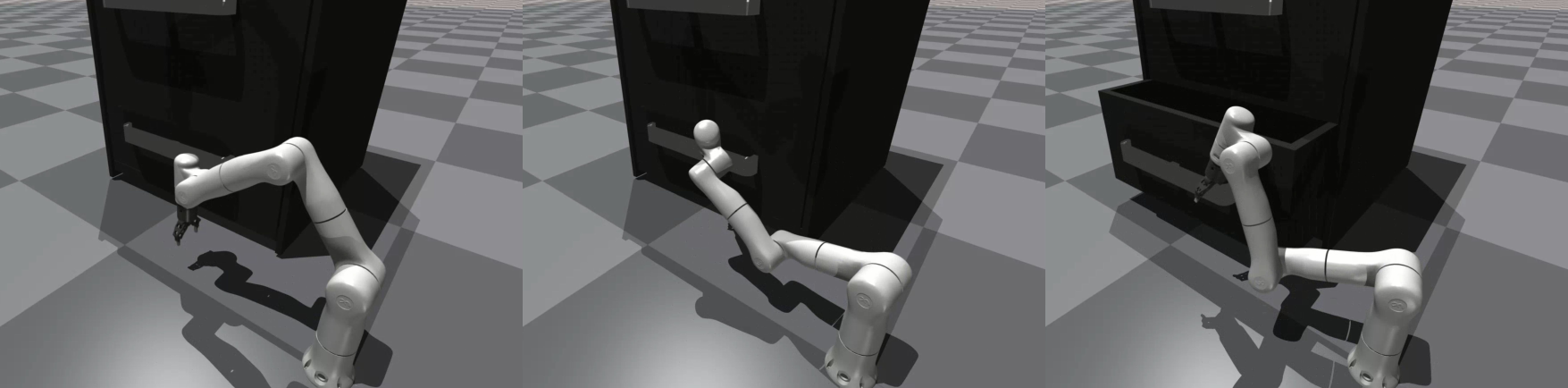}\label{fig:showd}
    \centering
    \end{minipage}}
\subfigure{
 \begin{minipage}[t]{0.98\linewidth}
    \includegraphics[width=\textwidth]{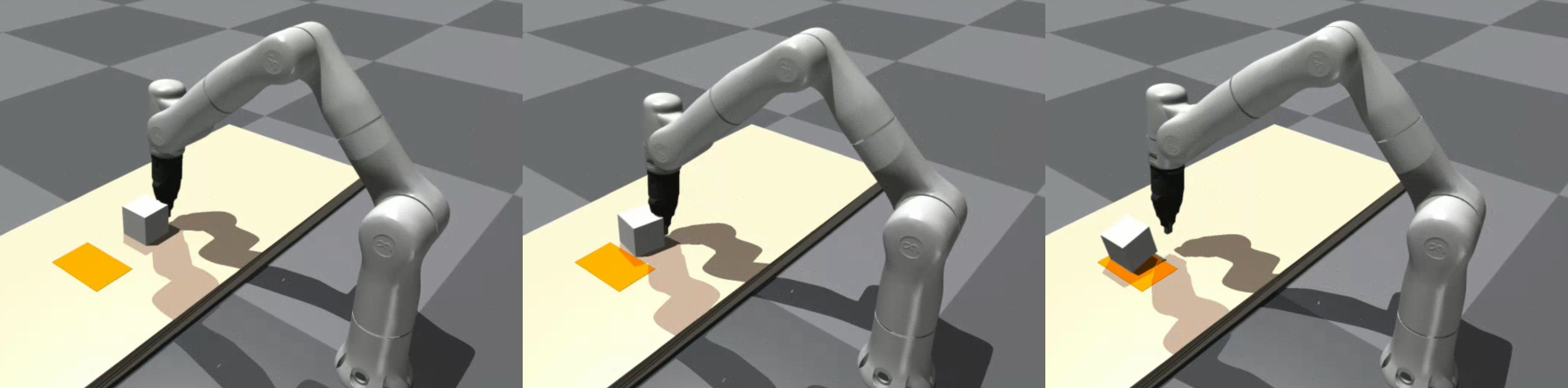}\label{fig:showh}
    \centering
    \end{minipage}}}
\resizebox{1\linewidth}{!}{\subfigure{
 \begin{minipage}[t]{0.98\linewidth}
    \includegraphics[width=\textwidth]{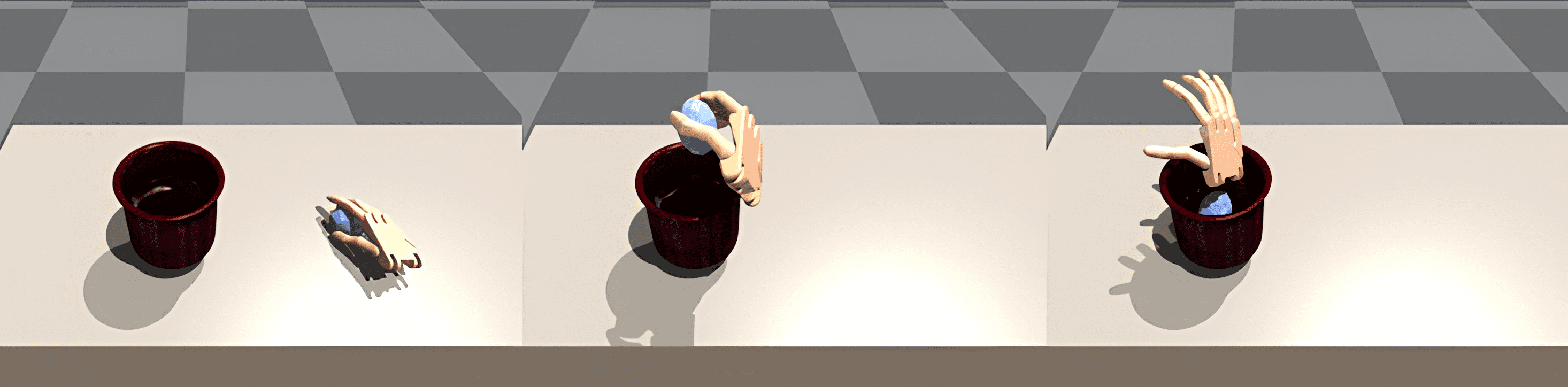}\label{fig:showp}
    \centering
    \end{minipage}}
\subfigure{
 \begin{minipage}[t]{0.98\linewidth}
    \includegraphics[width=\textwidth]{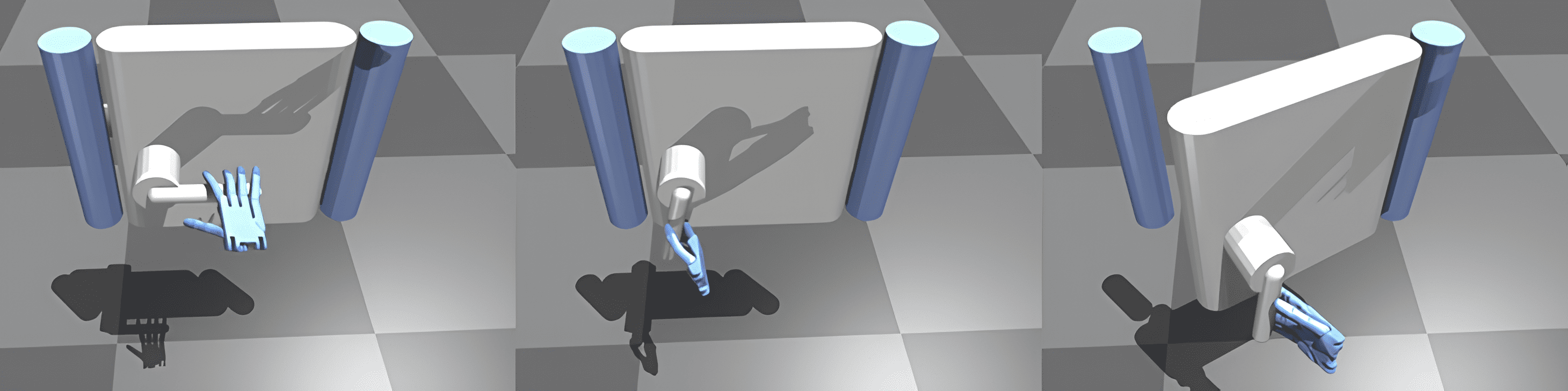}\label{fig:showd}
    \centering
    \end{minipage}}
\subfigure{
 \begin{minipage}[t]{0.98\linewidth}
    \includegraphics[width=\textwidth]{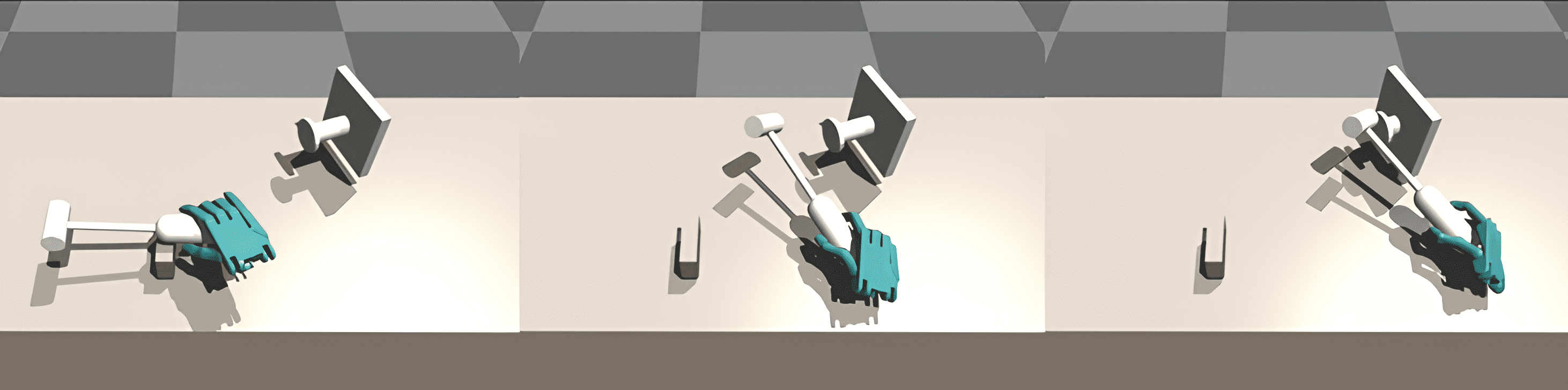}\label{fig:showh}
    \centering
    \end{minipage}}}
\caption{\textbf{Simulation tasks overview.} Here are six task settings and their task flow for Pick-and-Place \emph{(left)}, Articulated-Manipulation \emph{(middle)}, Gripper-Push \emph{(upper-right)} and Dexterous-Tool-Use \emph{(bottom-right)}.}
\label{fig:tasks}
\vspace{-10px}
\end{figure*}

\subsection{Teleoperation System.}
\label{sec:vbts}
Our system initially captures the raw sensory signal $\mathcal{I}$. Human hand pose $\mathcal{P}^h$  can be obtained from the captured signal using off-the-shelf 3D hand pose estimation~\cite{lv2021handtailor,rong2021frankmocap,weichert2013analysis}.
The pose $\mathcal{P}^h$ consists of the positions of the human hand's keypoints. Then, employing an inverse kinematic function $f_{\textit{IK}}$, we compute the action of the robot $a \in \mathbb{R}^{m}$, such that $a = f_{\textit{IK}}(\mathcal{P}^h_t,\mathcal{P}^h_{t+1})$, where it is calculated upon the change in the hand pose.
Given this teleoperation system, the human operator will move the hand to produce a sequence of hand poses $\{\mathcal{P}^h_i\}_{i=0}^T$ to teleoperate the robot with an action sequence $\{a_i\}_{i=0}^T$ to achieve the task $\mathcal{T}$. The 
human collected demonstration trajectory $\{(s_i,a_i)\}_{i=0}^{T}$, where $s \in \mathbb{R}^{n}$ is the robot state, could be used for downstream tasks.

\subsection{Diffusion-Model-Based Assistive Agent.} \label{sec:dp} 

After collecting data via teleoperation mentioned in Sec.~\ref{sec:vbts}, we train a diffusion-model-based assistive agent to learn how to assist humans in collecting data in a shared control manner.

At an abstract level, the diffusion-model-based assist agent, noted as $f(\cdot | \cdot)$, is provided with the state $s$, denoising step number $k$, and a noise action $a^k$, which could be an imperfect action gathered from the teleoperation system or sampled from a Gaussian distribution, to predict the desired action
\begin{equation}
\label{eqn:inference}
    a = f(a^k | s, k).
\end{equation}

\begin{figure}[t!]
    \centering
     \includegraphics[width=0.8\linewidth]{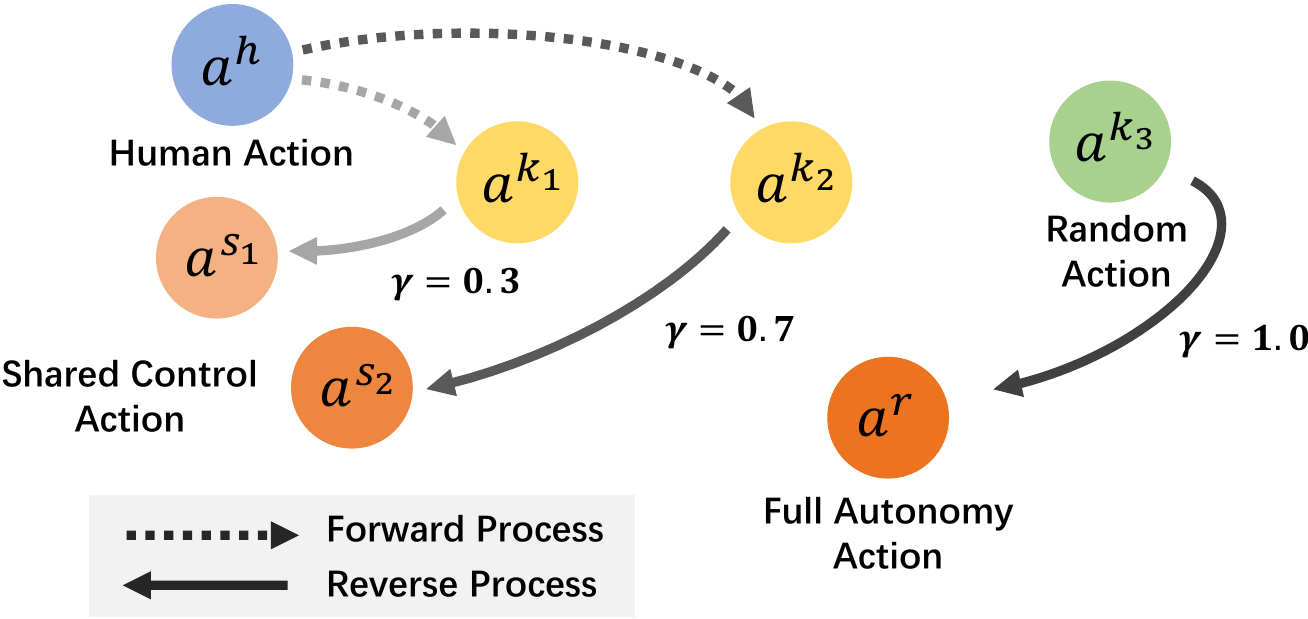}
    \caption{\textbf{Diffusion based shared control.} To achieve shared control between the human and agent, we blend the action from the human operator $a^h$ using the forward and reverse process. The parameter $\gamma$ governs the control ratio, where a lower $\gamma$ results in the action better aligning with the human operator's intention. In contrast, a higher $\gamma$ allows the learned agent to exert more influence over the blended action.}\label{fig:share}
\vspace{-15px}
\end{figure}


During data collection, the proposed system offers the option to control the robot in a shared control mode rather than directly applying the collected action $a^h$ from the teleoperation system. This leads to a reduced human workload during the data collection process.
The classical shared autonomy method is achieved through the equation~\cite{dragan2013policy}: 
\begin{equation}
a^s = \gamma a^h + (1 - \gamma)a^r, 
\label{eqn:sharecontrol}
\end{equation}
where $a^r$ is generated by the learned agent. 
However, considering that the agent operates as a diffusion policy (Fig.~\ref{fig:share}), we blend the action from the human with the forward and reverse processes. Given action $a^h$, a forward process diffuses the action as follows: $a^k = a^h + \epsilon^k.$
Subsequently, a reverse process denoises the action $a^k$:
\begin{equation}
a^s = f(a^k | s, k).
\label{eqn:diff_sharecontrol}
\end{equation}

By applying action $a$, the control of the robot is shared between the human and the diffusion-model-based assistive agent. We can adjust the control ratio $\gamma = k/K$ between the human operator and the diffusion-model-based assistive agent by varying $k$. 
When $\gamma = 0$, the action $a^s$ represents the teleoperation action $a^h$, which is the dexterous robot directly controlled by a human operator. As $\gamma$ approaches $1.0$, the action $a^s$ transitions to full autonomy $a^r$. 
A higher $\gamma$ value indicates a higher level of autonomy, allowing the learning-based agent more control rights to stabilize and direct the dexterous hand.

\begin{algorithm}[t]
	\caption{Overall Process}
	\label{alg:overall}
	\begin{algorithmic}[1]
		\Require The human operator $\mathcal{H}$;  
		\Ensure The collected dataset $\mathcal{D}$; assistive agent $f$; control ratio $\gamma$; 
		\State Initialization: $\mathcal{D} \gets \emptyset, \gamma \gets 0$; 
		\While{$|\mathcal{D}|$ is small} \Comment{not enough data is collected}
            \State $\mathcal{H}$ collects data $d$ under $f$'s help; 
            \Comment{see \ref{algo:control_adjust} for control ratio adjustment} 
            \If{$d$ is valid} 
                \State $\mathcal{D} \gets \{d\} \cup \mathcal{D}$;
            \EndIf
            \State Finetune $f$ with $\mathcal{D}$;
        \EndWhile
		\State \Return $\mathcal{D}$ and $f$;
	\end{algorithmic}
\end{algorithm}

\subsection{Integrating Data Collection and Manipulation Learning.} \label{sec:process} 
In this section, we show how to integrate data collection and manipulation learning into a unified framework that progressively reduces human effort and enhances robot autonomy.

\subsubsection{Detailed Algorithm Explanation}
We outline the overall process in Algo.~\ref{alg:overall}. The assistive agent is trained in three steps as follows:

\textit{Step 1.} Initially, we collect a dataset for pre-training agent $f$ under full manual control by human operators, \textit{i.e.}, with the control ratio $\gamma =0$.

\textit{Step 2.} Given the initial dataset, we train a relatively low performance assistive agent to aid in further data collection. The training process has been formulated in Eq.~\ref{eqn:loss} and Eq.~\ref{eqn:inference}, where a neural network $\epsilon_\theta$ is trained to predict noise $\epsilon$ out of the noisy action $a^k$.

\textit{Step 3.} The trained agent assists in a second data collection round, aiming for higher efficiency and success. We then refine the agent using data from both rounds to enhance its performance. This cycle repeats until the agent achieves full autonomy and the required data volume is collected.

\subsubsection{Control Ratio Adjustment}
\label{algo:control_adjust}
For each data collection, we offer users two options to adjust the control ratio $\gamma$: (1) Users can empirically adjust $\gamma$ based on their needs. (2) Alternatively, set $\gamma=\frac 1 2 (1+\cos\theta)$, where $\theta$ is obtained by calculating the dot product of the previous timestep's human action $a^h$ and shared action $a^s$. This assesses alignment, increasing $\gamma$ for positive alignment to enhance agent control, and reducing it for misalignment to increase user control.

After obtaining the control ratio $\gamma$, we calculate the shared action $a^s$, using the human operator's action $a^h$ as input, as defined in Eq.~\ref{eqn:diff_sharecontrol} (shown in Fig.~\ref{fig:share}).

\section{EXPERIMENTS}
In this section, we introduce the settings for both real world tasks and simulation tasks, along with the experimental results and data validation. Due to page limitations, we have included some detailed information such as training details and ablation study results on our webpage.

\subsection{Tasks.} 
\label{task_settings}
We adopt six multi-stage manipulation tasks (Fig.~\ref{fig:tasks}). 
\emph{Pick-and-Place} aims at picking an object on the table and placing it into a container. \emph{Articulated-Manipulation}'s objective for the dexterous hand is to grasp and unscrew a door handle to open it, while for the gripper, it is to grab a drawer handle and pull the drawer open. 
\emph{Push-cube} requires the robot to push the cube to the target position. 
\emph{Tool-Use} aims at picking a hammer and using it to drive a nail into a board. 

\begin{table*}[t]
\setlength{\abovecaptionskip}{0cm}
\setlength{\belowcaptionskip}{0cm}
\setlength{\extrarowheight}{0pt} 
\renewcommand{\arraystretch}{0.8} 
\begin{center}
\resizebox{0.99\linewidth}{!}{\begin{tabular}{cc|ccc|ccc|ccc}
\toprule
 && \multicolumn{3}{c|}{\emph{Pick-and-Place}} & \multicolumn{3}{c|}{\emph{Door-Open}} & \multicolumn{3}{c}{\emph{Tool-Use}} \\
 && \emph{Success} & \emph{Horizon} & \emph{Collection} & \emph{Success} & \emph{Horizon} & \emph{Collection} & \emph{Success} & \emph{Horizon} & \emph{Collection} \\
 && \emph{Rate} $\uparrow$ & \emph{Length} $\downarrow$& \emph{Speed} $\uparrow$ & \emph{Rate} $\uparrow$ & \emph{Length} 
 $\downarrow$ & \emph{Speed} $\uparrow$ & \emph{Rate} $\uparrow$ & \emph{Length} $\downarrow$ & \emph{Speed} $\uparrow$ \\
\midrule\midrule
 \emph{Group1}&\emph{w/ Ours} & \textbf{86.96}& \textbf{219.01}& \textbf{320} & \textbf{87.11} & \textbf{142.29} & \textbf{460} & \textbf{66.50} & \textbf{232.17} & \textbf{200}\\
 &\emph{w/o Ours} & 51.53& 378.49& 176& 62.49& 258.27& 252& 42.38& 487.95&129\\
\midrule\midrule
 \emph{Group2}& \emph{w/ Ours}& \textbf{94.06} & \textbf{214.16} & \textbf{324} & \textbf{80.29}& \textbf{134.16} & \textbf{424} & \textbf{55.55}& \textbf{275.71}&\textbf{172}\\
 & \emph{w/o Ours} & 45.42& 471.48& 120& 53.45& 317.21& 176& 34.47& 511.03&124\\
 \bottomrule
\end{tabular}}
\end{center}
\caption{User studies on three dexterous hand tasks.}
\label{tab:userstudy}
\vspace{-15px}
\end{table*}

\begin{figure*}[t]
\centering
\resizebox{1\linewidth}{!}{\subfigure{
 \begin{minipage}[t]{0.98\linewidth}
    \includegraphics[width=\textwidth]{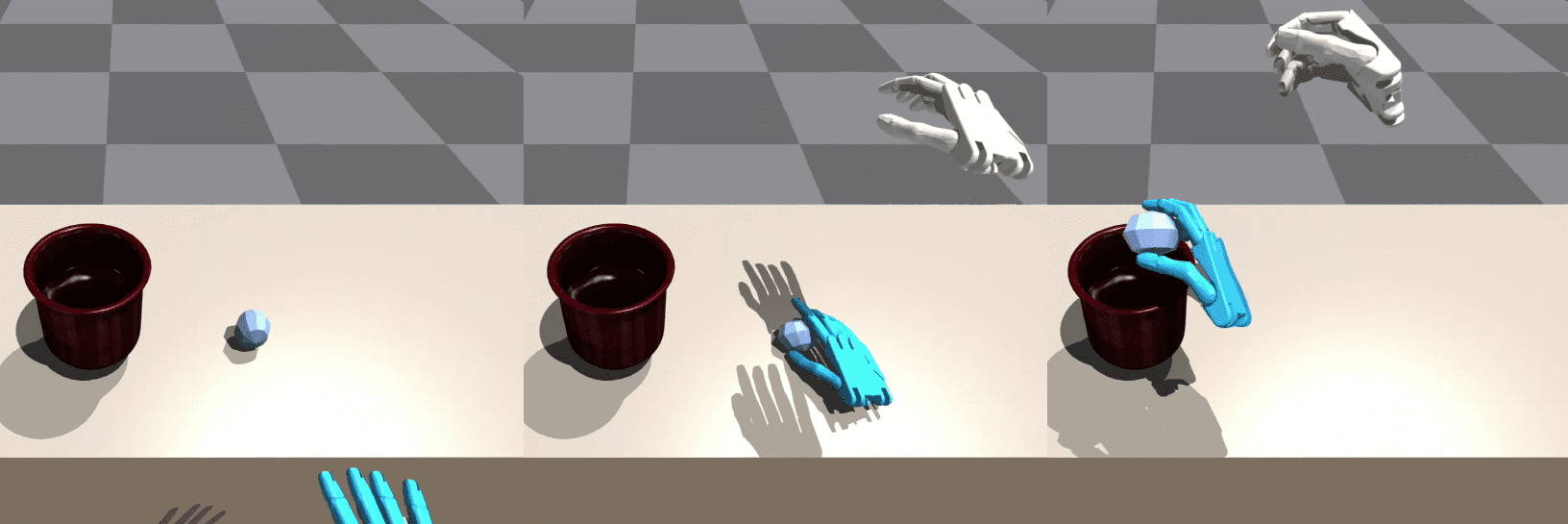}\label{fig:showp}
    \centering
    \end{minipage}}
\subfigure{
 \begin{minipage}[t]{0.98\linewidth}
    \includegraphics[width=\textwidth]{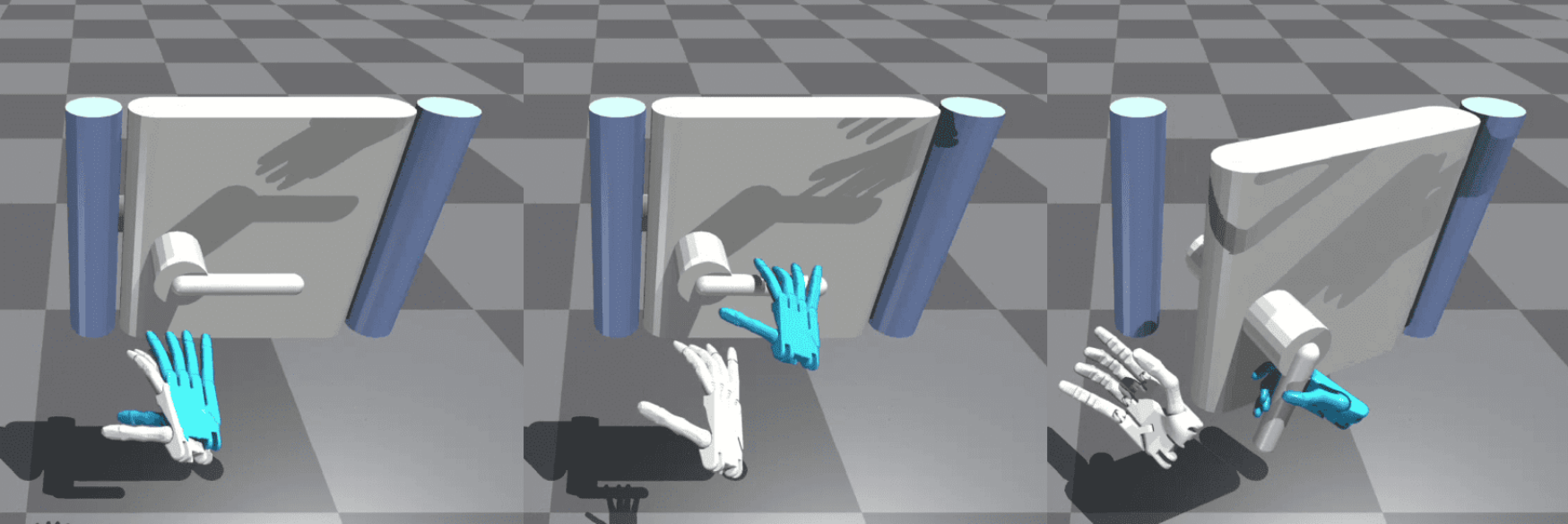}\label{fig:showd}
    \centering
    \end{minipage}}
\subfigure{
 \begin{minipage}[t]{0.98\linewidth}
    \includegraphics[width=\textwidth]{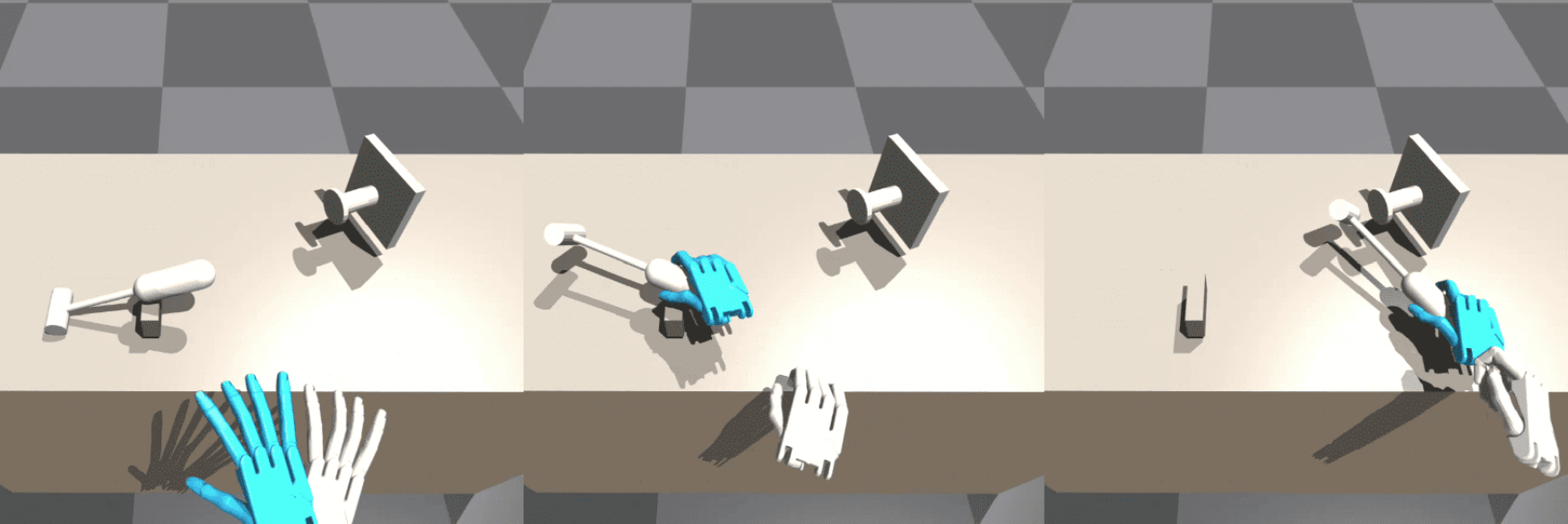}
    \centering
    \end{minipage}}}
\caption{\textbf{Shared control process overview.} The white one is the hand controlled purely by the \textit{human operator}, while the cyan one is under \textit{shared control} between the human and the assistive agent.}
\label{fig:viz}
\vspace{-15px}
\end{figure*}

\subsection{Efficiency of Data Collection.} 
\label{user_study} 
Our proposed system leverages shared control between human operators and learned agents to enhance the efficiency of data collection. To learn how the assistant agent could improve the data collection process, we conducted a user study.

In the user study, 10 human operators participate, collecting data under two modes: one where control is shared between the operator and the learned agent (\emph{w/ Ours}), and the other where control is directly by the operator alone (\emph{w/o Ours}). Each participant is instructed to collect as much data as possible within three minutes under two different modes for three dexterous hand tasks. Three metrics are evaluated: \emph{Success Rate} (Percent) indicates the percentage of attempts where data collection was successful. \emph{Horizon Length} (Steps per Sample) measures the length of each collected trajectory, with a lower horizon length indicating smoother data collection. \emph{Collection Speed} (Samples per Hour) refers to the number of successful trajectories that can be collected in one hour.

In Tab.~\ref{tab:userstudy}, by sharing control between humans and learned agents, our system shows improvements in both success rate and collection speed, while the average horizon length of the collected trajectories is reduced. This suggests that our system enhances the efficiency of data collection by facilitating a process that is easier to succeed, faster, and more fluid in terms of trajectory smoothness. To ensure the fairness of the experiment wasn’t compromised, we equally divided the user group into two parts, \emph{Group 1} first collected data directly by themselves \emph{(w/o Ours)} and then collected data with an assistive agent \emph{(w/ Ours)}, while the \emph{Group 2} reversed the order, first \emph{(w/ Ours)} mode and then \emph{(w/o Ours)} mode.




\subsection{Quantitative Evaluation.} To gain deeper insight into how the learned agent assists the human operator, we visualize several keyframes from the data collection process of three dexterous hand tasks. From Fig.~\ref{fig:viz}, it is evident that human operators are not required to provide too precise control with the assistive agent facilitating shared control over the dexterous hand. Instead, they only need to convey high-level intentions, such as the direction of hand movement or finger grasp motions. In multi-stage tasks, like picking up a hammer and then using it to drive a nail, operators only need to provide a \textit{trigger action} to guide the agent to transition from one sub-stage to the next. As a result, less effort and attention are required, making the data collection easier to execute successfully and speeding it up.

When the learned agent shares control with users, the system effectively corrects imperfect human control signals to accomplish specific tasks. 
Given the challenge of directly measuring the level of imperfection in user signals and the correction ability of our system, we simulate human input using a baseline agent trained with Behavior Cloning (BC) as a proxy for user control.

\begin{figure*}[t]
\setlength{\abovecaptionskip}{0cm}
\setlength{\belowcaptionskip}{0cm}
\centering
\subfigure[\emph{Pick-and-Place}]{
 \begin{minipage}[t]{0.31\linewidth}
    \includegraphics[width=\textwidth]{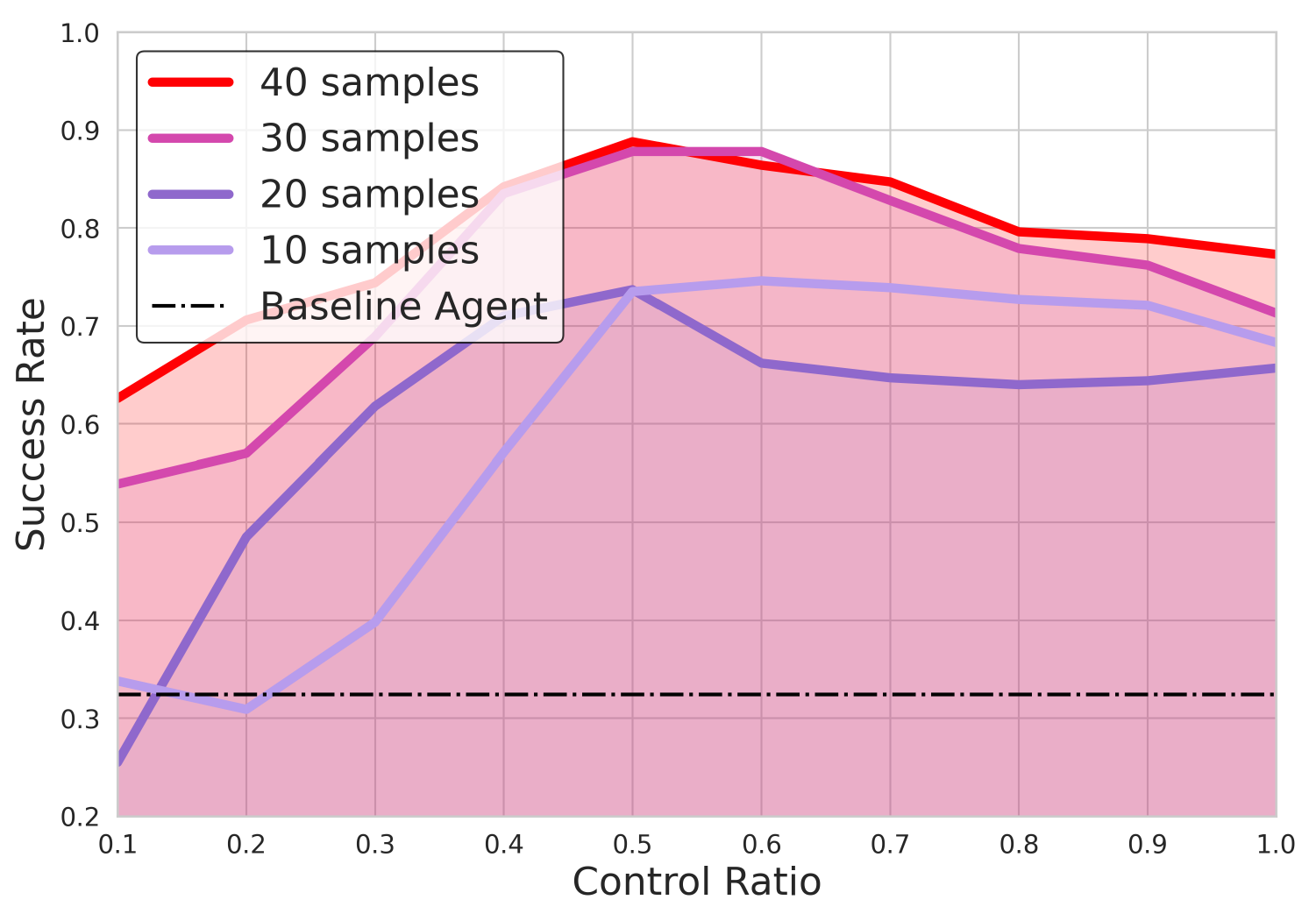}
    \centering
    \end{minipage}}
\subfigure[\emph{Articulated}]{
 \begin{minipage}[t]{0.31\linewidth}
    \includegraphics[width=\textwidth]{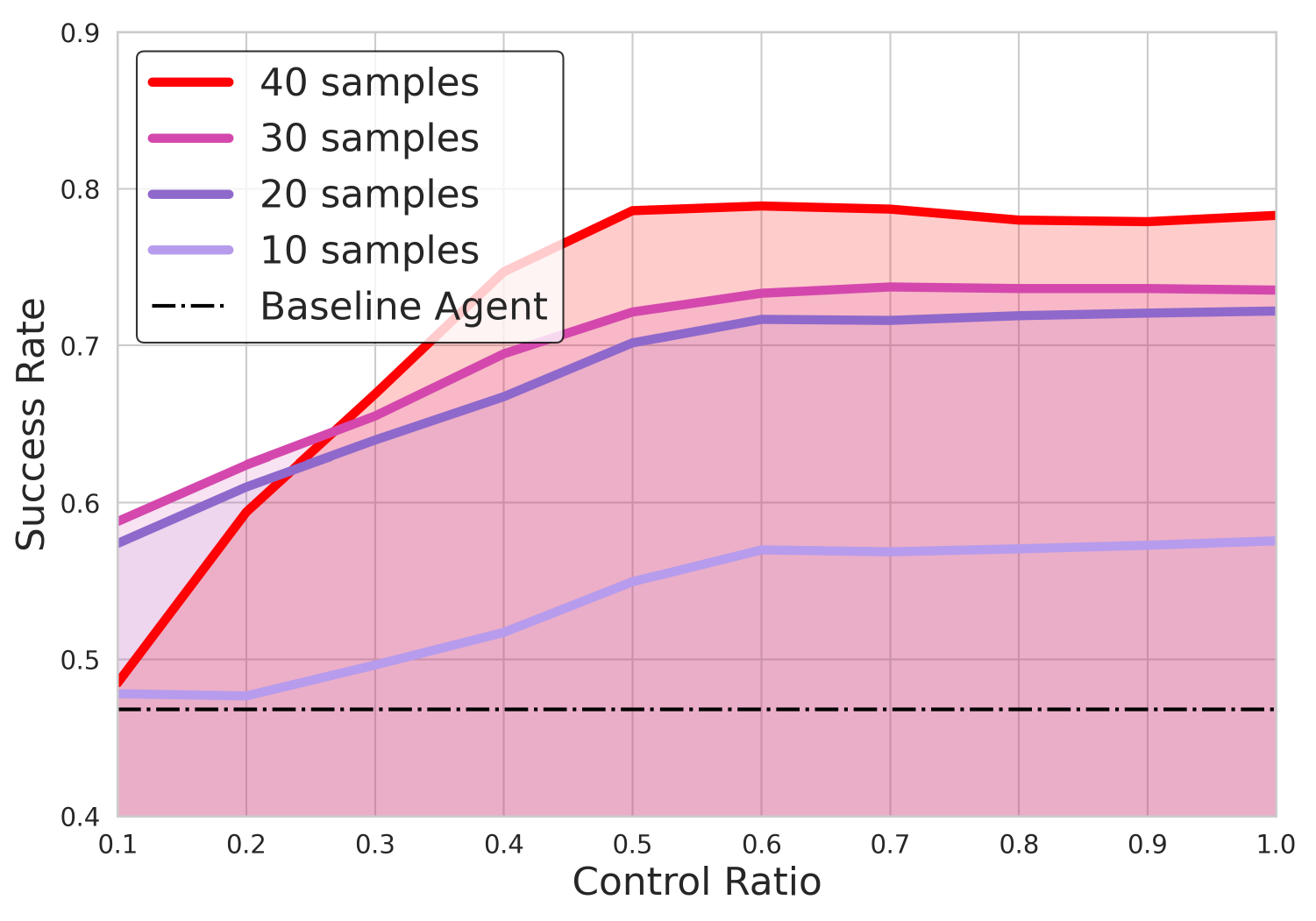}
    \centering
    \end{minipage}}
\subfigure[\emph{Tool-Use}]{
 \begin{minipage}[t]{0.31\linewidth}
    \includegraphics[width=\textwidth]{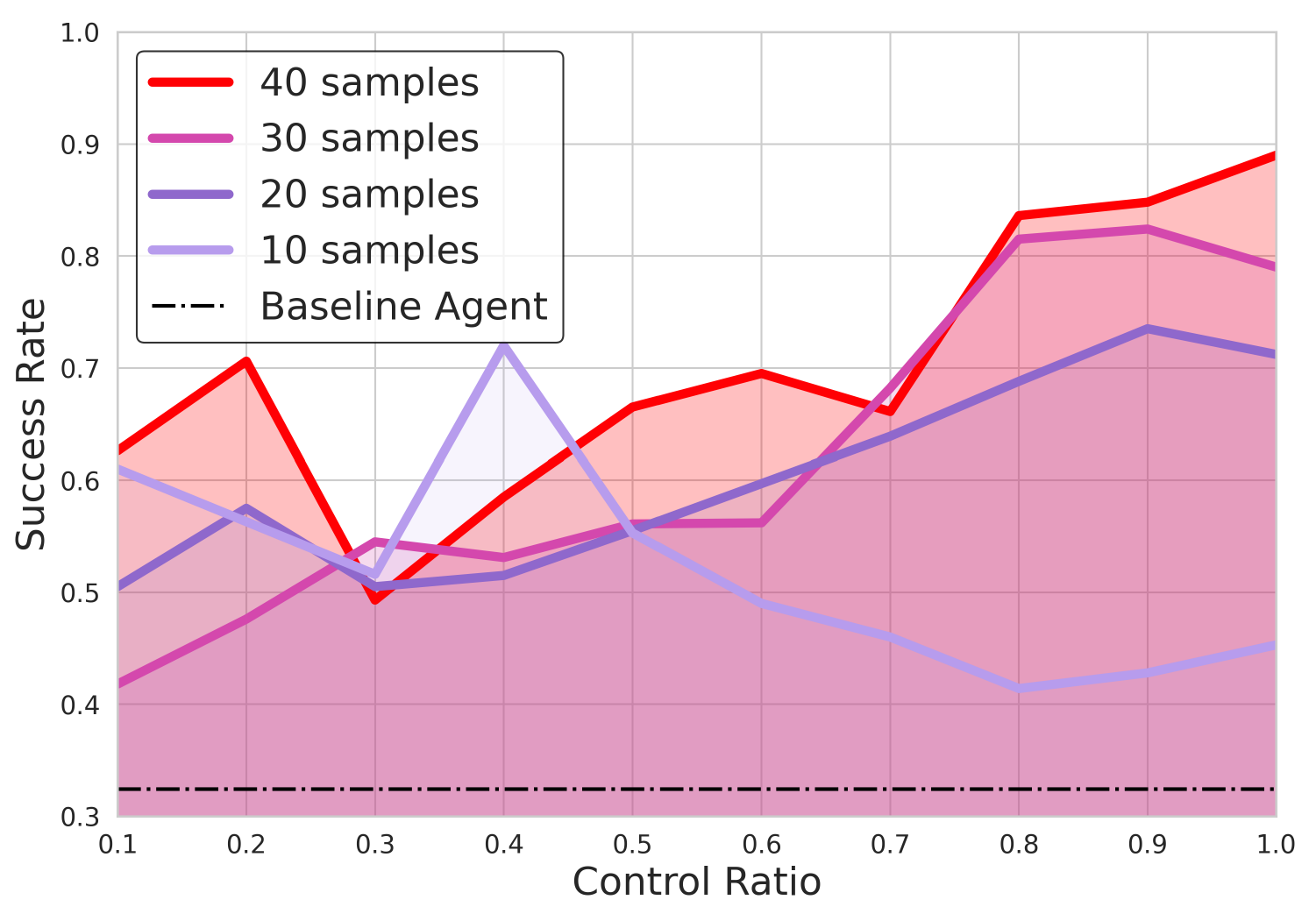}
    \centering
    \end{minipage}}
\caption{\textbf{Agent performance over time.} The \emph{x-axis} represents the control ratio $\gamma$ and the \emph{y-axis} represents the success rate. We train a simulated operator to evaluate our system, it shows that even with limited data, the learned assist agent can improve the success rate of data collection to improve the efficiency. 
With the data accumulated, the performance of the learned agent keeps rising. Moreover, the learned agent could be transitioned to a full autonomy agent ($\gamma=1.0$).}
\label{fig:simulateduser}
\vspace{-15px}
\end{figure*}

In Fig.~\ref{fig:simulateduser}, additional data collected under our framework effectively contributes to training improvements. The graph illustrates that with limited data availability, the agent can assist the simulated operator more effectively. As the agent gains access to and trains on more data, its ability to correct actions improves. These results indicate that our system gradually reduces the demand for the operator's attention and effort, thereby enhancing the overall efficiency of data collection process.

Furthermore, once sufficient data is collected and the assistive agent is trained, it can transition into full autonomy mode by setting $\gamma = 1$ and denoising actions from the noise sampled from Gaussian distributions. Across three different dexterous manipulation tasks, we can achieve success rates of 0.76, 0.78, and 0.89, indicating that the assistive agent can effectively transform into an automated dexterous manipulation agent.

From our experiments, we have observed that the assistive agent significantly aids human operators in managing fine control, especially in scenarios where accurate observation by humans is challenging. 
For instance, tasks such as grasping an egg or moving a hammer present visual challenges. It can be difficult to visually confirm whether the egg is securely grasped or if there's a risk of it being dropped. This uncertainty makes it hard for human operators to react promptly to sudden changes. However, within our proposed joint learning framework, human operators are primarily required to focus on high-level intentions and task planning during data collection, while the assistive agent manages the detailed low-level actions. This division of labor significantly reduces the burden on human operators by clearly separating strategic planning from execution tasks, streamlining the collaboration between humans and machines.

\begin{table}[t!]
\setlength{\abovecaptionskip}{0cm}
\setlength{\belowcaptionskip}{0cm}
\begin{center}
\resizebox{1\linewidth}{!}{\begin{tabular}{c|cc|cc|cc}
\toprule
Dexterous& \multicolumn{2}{c|}{\emph{Pick-and-Place}} & \multicolumn{2}{c|}{\emph{Articulated-Manipulation}} & \multicolumn{2}{c}{\emph{Tool-Use}}\\
Hand& \emph{40$\mathcal H$} & \emph{10$\mathcal H$ + 30$\mathcal S$} & \emph{40$\mathcal H$} & \emph{10$\mathcal H$ + 30$\mathcal S$} & \emph{40$\mathcal H$}& \emph{10$\mathcal H$ + 30$\mathcal S$}  \\
\midrule
\emph{BC} & 0.30 & \textbf{0.50} & 0.22 & \textbf{0.57} & 0.39& \textbf{0.40}  \\
\emph{BC-RNN} & 0.54 & \textbf{0.67} & 0.47 & \textbf{0.50} & \textbf{0.27}& 0.25  \\
\emph{DP} & 0.73 & \textbf{0.76} & 0.77 & \textbf{0.78} & 0.88& \textbf{0.89}  \\
\bottomrule
\toprule
Parallel& \multicolumn{2}{c|}{\emph{Pick-and-Place}} & \multicolumn{2}{c|}{\emph{Articulated-Manipulation}} & \multicolumn{2}{c}{\emph{Push-cube}}\\
 Gripper& \emph{40$\mathcal H$}& \emph{10$\mathcal H$ + 30$\mathcal S$}& \emph{40$\mathcal H$}& \emph{10$\mathcal H$ + 30$\mathcal S$}& \emph{40$\mathcal H$}& \emph{10$\mathcal H$ + 30$\mathcal S$}  \\
\midrule
\emph{BC} & 0.42&  \textbf{0.44} & 0.35& \textbf{0.37} & \textbf{0.88}& 0.85\\
\emph{BC-RNN} & \textbf{0.39}& 0.36& 0.71 & \textbf{0.73} & 0.59& \textbf{0.67}  \\
\emph{DP} & 0.51& \textbf{0.60} & 0.42& \textbf{0.67} & \textbf{0.83}& 0.82  \\
\bottomrule
\end{tabular}}
\end{center}
\caption{Data quality on downstream tasks. }
\label{tab:downstream}
\vspace{-10px}
\end{table}

\begin{table}
\setlength{\abovecaptionskip}{0cm}
\setlength{\belowcaptionskip}{0cm}
  \begin{center}
  \renewcommand{\arraystretch}{0.7}
  \resizebox{0.8\linewidth}{!}{
	\begin{tabular}{c|cc|cc}
        \toprule
        & \multicolumn{2}{c|}{\emph{Dexterous Tool-Use}} & \multicolumn{2}{c}{\emph{Gripper Push-cube}}\\
	& \emph{BC} & \emph{DP} & \emph{BC} & \emph{DP}\\
	\midrule
	10$\mathcal H$ & 0.29 & 0.45 & 0.23& 0.42\\
	\midrule
	10$\mathcal H$ + 10$\mathcal H$& 0.28 & 0.67 & 0.37& 0.78\\
	10$\mathcal H$ + 20$\mathcal H$& 0.28 & 0.82 & 0.51& 0.67\\
	10$\mathcal H$ + 30$\mathcal H$& 0.39 & 0.88 & 0.88& 0.83\\
	\midrule
	10$\mathcal H$ + 10$\mathcal S$ & 0.31 & 0.71 & 0.33& 0.81\\
	10$\mathcal H$ + 20$\mathcal S$ & 0.30 & 0.79 & 0.61& 0.62\\
	10$\mathcal H$ + 30$\mathcal S$ & 0.40 & 0.89 & 0.85& 0.82\\
	\bottomrule
	\end{tabular}}
  \end{center}
  \caption{Agent performance under increasing data.}
  \label{tab:abla_bc}
\vspace{-20px}
\end{table}

\subsection{Data Quality on Downstream Task.}
In this section, we illustrate that collecting data under shared control does not compromise the quality of the data. We gather dexterous hand and gripper manipulation demonstrations via the proposed system in two modes: fully controlling the robots by a human (\emph{$\mathcal H$}) and sharing control (\emph{$\mathcal S$}) between the human operator and the learned assistive agent. And utilize these data to train different kinds of agents, like BC, BC-RNN~\cite{mandlekar2021matters}, and Diffusion Policy~(DP)~\cite{chi2023diffusion}. 


In Tab.~\ref{tab:downstream}, compared to directly collecting human demonstrations from the expert human operator, who can achieve success rates and efficiency comparable to those with agent assistance, the data collected by sharing control between the human and the assistive agent can achieve comparable or even surprisingly better results with BC and BC-RNN. Their results are comparable with DP, possibly as DP can better fit the tasks, which is in line with \cite{chi2023diffusion}.


  
In Tab.~\ref{tab:abla_bc}, we compare the effects of using different sets of data to train BC and DP. We can find that utilizing more data collected under the shared control mode leads to comparable performance on the tool-use and push-cube tasks. 
This verifies that the new data contributes significantly to policy learning and can achieve a similar effect compared to the data from human experts but at a much lower cost.
These results indicate that the data collected under the proposed paradigm have sufficient quality and efficiency for downstream tasks.

\subsection{Real World Experiment and User Feedback.}
\label{realworld}
\begin{figure}[t]
\setlength{\abovecaptionskip}{0cm}
\setlength{\belowcaptionskip}{0cm}
    \centering
    \includegraphics[width=\linewidth]{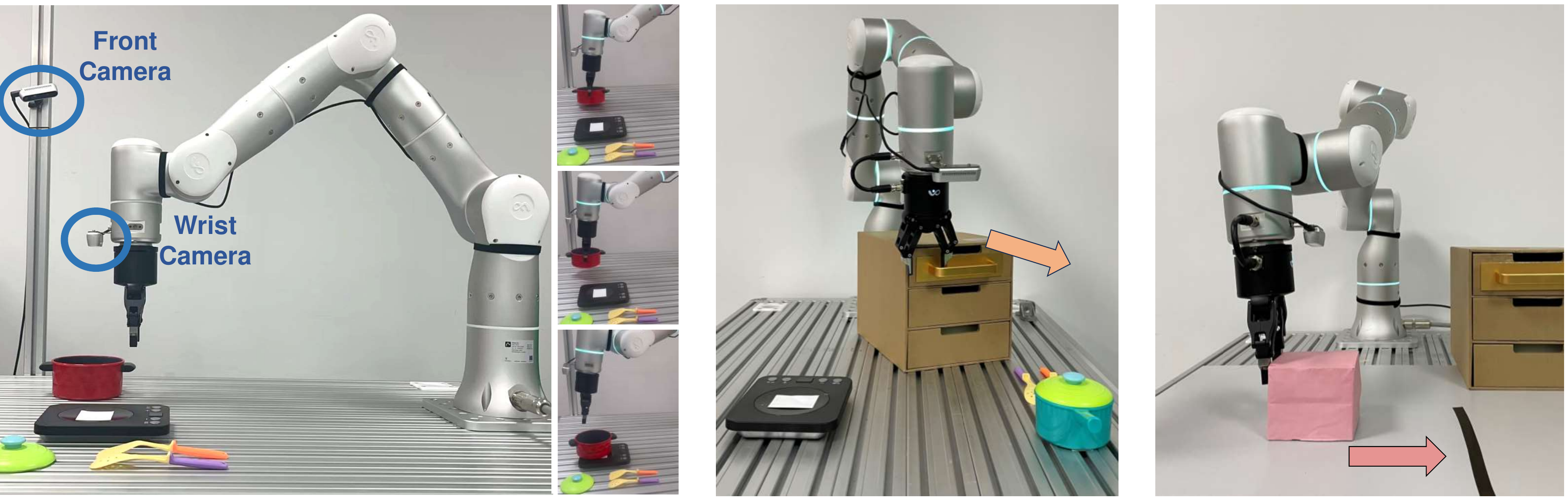}
\caption{\textbf{Real world setting.} 1. \textit{Pick-and-Place}: use the gripper to pick the red pot up and place it onto the black induction cooker. 2. \textit{Articulated-Manipulation}: use the gripper to open the drawer. 3. \textit{Push-cube}: use the gripper to push the cube across the black line.}
    \label{fig:real_world_scene}
    \vspace{-10px}
\end{figure}

To better evaluate our system, we further conduct real-world experiments. Three tasks are adopted: Pick-and-Place, Articulated-Manipulation, and Push-cube in Fig.~\ref{fig:real_world_scene}.
Following the same rules as Sec.~\ref{user_study}, four human volunteers are invited to participate in the user study to collect data under two modes: one where control is shared between the human operator and the learned agent (\emph{w/ Ours}), and the other where control is directly by the human operator alone (\emph{w/o Ours}). 
Our proposed system achieves significant improvements in success rate and collection speed by sharing control between human operators and learned agents, as demonstrated in Tab.~\ref{tab:real_world}. Additionally, data gathered under our proposed joint learning shared control mode yield performance on the three tasks that are comparable to those pure human datasets using BC and DP, further substantiated by the results presented in Tab.~\ref{tab:realworld_exec}.

\begin{table}[t!]
\setlength{\abovecaptionskip}{0cm}
\setlength{\belowcaptionskip}{0cm}
\begin{center}
	\resizebox{0.9\linewidth}{!}{\begin{tabular}{c|c|c|c}
	\toprule
	& \emph{Success Rate $\uparrow$}& \emph{Horizon Length $\downarrow$}&\emph{Collection Speed $\uparrow$}\\
	\midrule
	\emph{w/ Ours}& \textbf{0.79} & \textbf{18.72} & \textbf{151}\\
	\midrule
	\emph{w/o Ours}& 0.70 & 21.54 & 121 \\
    \bottomrule
	\end{tabular}}
	\end{center}
\caption{Real world gripper Pick-and-Place task user study.}
\label{tab:real_world}
\vspace{-20px}
\end{table}

\begin{table}[t!]

\setlength{\abovecaptionskip}{0cm}
\setlength{\belowcaptionskip}{0cm}
  \begin{center}
  
  \resizebox{1\linewidth}{!}{\begin{tabular}{c|cc|cc|cc}
  \toprule
  & \multicolumn{2}{c|}{\emph{Pick-and-Place}}  & \multicolumn{2}{c|}{\emph{Articulated-Manipulation}}  & \multicolumn{2}{c}{\emph{Push-cube}} \\
  &\emph{40$\mathcal H$} & \emph{20$\mathcal H$ + 20$\mathcal S$}  &\emph{30$\mathcal H$} & \emph{10$\mathcal H$ + 20$\mathcal S$}  &\emph{20$\mathcal H$} & \emph{10$\mathcal H$ + 10$\mathcal S$}\\
  \midrule
  \emph{BC} & 13 / 20 & 14 / 20 & 18 / 20 & 19 / 20 & 15 / 20 & 15 / 20 \\
  \emph{DP} & 11 / 20 & 12 / 20 & 16 / 20 & 12 / 20 & 15 / 20 & 13 / 20 \\
  \bottomrule
  
  \end{tabular}}
  \end{center}
  \captionof{table}{Real world gripper experiments of data quality.}
  \label{tab:realworld_exec}
\vspace{-5px}
\end{table}

\begin{table}[t!]
\setlength{\abovecaptionskip}{0cm}
\setlength{\belowcaptionskip}{0cm}
\begin{center}
\resizebox{1\linewidth}{!}{\begin{tabular}{ll}
\hline\hline
\textit{\textbf{Satisfaction}: $\alpha = 0.769$} &  \\
\textit{1. It is fun to use.} &  \\
\textit{2. It works the way I want it to work.} &  \\
\textit{3. It is wonderful.} &  \\
\textit{4. It helps me be more effective.} &  \\
\textit{5. It is flexible.} &  \\
\midrule
\textit{\textbf{User-Friendly}: $\alpha = 0.852$} &  \\
\textit{6. It is simple to use.} &  \\
\textit{7. It is effortless.} &  \\
\textit{8. I can use it without written instructions.} &  \\
\textit{9. I do not notice any inconsistencies as I use it.} &  \\
\hline\hline
\end{tabular}}
\end{center}
\captionof{table}{Subjective Measures.}
\label{tab:user_question}
\vspace{-20px}
\end{table}



We have developed a questionnaire comprising shown in Tab.~\ref{tab:user_question} to capture various dimensions of user experience and ergonomics, and we invited 10 volunteers to rate our system based on their feedback.

This questionnaire assesses ease of use and overall satisfaction. The reliability of our questionnaire is supported by strong Cronbach's alpha values: $\alpha = 0.769$ for the satisfaction section and $\alpha = 0.852$ for the user-friendly section, indicating internal consistency.

\section{CONCLUSION}

In this paper, we introduce a novel human-agent joint learning paradigm that enables simultaneous human demonstration collection and robot manipulation teaching. This approach allows the human operator to share control with a diffusion-model-based assistive agent within a vision-based teleoperation system to control multiple robot end-effectors such as grippers and dexterous hands. Given our paradigm, the human operator can reduce the effort spent on data collection and adjust the control ratio between the human and agent based on different scenarios. Our system offers a more efficient and flexible solution for data collection and robot manipulation learning via teleoperation. 

\bibliographystyle{IEEEtranN}
\bibliography{references}

\end{document}


\maketitle
\thispagestyle{empty}
\pagestyle{empty}

\section*{APPENDIX}

\section{Implementation details}

Here we lay down the details of the data collection, training, and testing process. More technical details are given here to illustrate our method and implementations better.

\subsection{Shadow Hand and Parallel Gripper Teleoperate System}

To adapt to Isaac Gym and our vision system, we made certain modifications to the XML file of the Shadow Hand. We followed \cite{wu2023graspgf,qin2023anyteleop,qin2022dexmv}, removed the entire arm part, and added six degrees of freedom to the base mount of the Shadow Hand. This allows it to move freely in the virtual environment without depending on a base. Similarly, to obtain the rigid body Jacobian matrices of the five fingertips of the Shadow Hand, we added a massless rigid body to the tips of all five fingers of the Shadow Hand. This facilitates direct inverse kinematics calculations for the entire finger.
In inverse kinematics (IK) calculations, we employed the Damped Least Squares (DLS) method~\cite{buss2005selectively,pechev2008inverse}, this approach helps to prevent instability issues when approaching singularity points. Additionally, the DLS method supports real-time applications because it can provide fast and stable solutions, which is particularly crucial for teleoperation systems. Focusing solely on the five fingertips and wrist is regarded as the most balanced approach between computational efficiency and the precision required for complex hand movements in real-time applications. The system operates on a computer with an RTX 4070 graphics card and a monitor. 

To mitigate the accumulation of errors, the process involves mapping hand motion from the real world into the virtual environment and then comparing each action with the action from the previous frame to calculate a delta action. The reason for calculating delta action is to identify and apply only the changes in movement from one frame to the next, rather than applying the absolute positions and orientations directly. This approach helps reduce the accumulation of errors that might occur due to discrepancies between the real-world movements and their representation in the simulated environment. By focusing on the changes (delta) rather than absolute values, the system can more accurately replicate the intended movements in the simulator, leading to more precise and consistent control of the shadow hand.

\subsection{Baselines}

In this section, we provide the implementation details for BC and BC-RNN models.  In Behavior Cloning (BC),  the objective is to minimize $\mathbb{E}_{(s,a)\sim\mathcal{D}}||\pi_\theta(s)-a||^2$. We use a 3-layer multi-layer perception (MLP) with a ReLU activation function.  All layers are fully connected layers with $128$ hidden dimensions with a learning rate of $2\cdot 10^{-3}$. We also use the AdamW \cite{loshchilov2017decoupled} to be the optimizer. The training epoch in dexterous tasks Pick-and-Place, Articulated-Manipulation, and Tool-Use is $60, 100, 100$ separately.

As for BC-RNN, we use an LSTM as the backbone network for BC-RNN \cite{mandlekar2021matters}, which we find a slight performance improvement compared to the vanilla RNN model. Following \cite{mandlekar2021matters}, during the training phase, a state-action sequence $\{(s_i, a_i), \cdots, (s_{i+T-1}, a_{i+T-1})\}$ of length $T$ is sampled from the dataset $\mathcal D$ and the network will predict the action sequence based on the states as its input.
During the inference phase $a_t,h_{t+1}={\pi_{\theta}}(s_t,h_t)$ where $h_t, h_{t+1}$ are the hidden states.  Here we set the learning rate to be $2\cdot 10^{-3}$, and the training epoch to be $60$.

\subsection{Diffusion-Model-Based Assistive Agent}
The assistive agent's noise prediction model $\boldsymbol \epsilon_\theta$'s backbone network is a 4-layer multi-layer perception (MLP) with a Softplus activation function. All layers are fully connected layers with 128 hidden dimensions. 
Moreover, we set the diffusion steps $K=50$, $\beta_{\min} = 10^{-4}, \beta_{\max}=0.1$ in Eq. 2 with Sigmoid scheduling and use Exponential Moving Average (EMA) to stabilize the training. The learning rate of $\boldsymbol \epsilon_\theta$ is $10^{-3}$.

\section{Experiment Setups}

\subsection{Tasks}
\textbf{\emph{Dexterous Hand Pick-and-Place}} aims at picking an object on the table and placing it into a container.
The observation space is 24 dimensions, including the dexterous robot hand state (18-dim), the object's position (3-dim), and the container's position (3-dim). The dexterous robot hand state is the position of each fingertip (15-dim) and the wrist position (3-dim). The action space is 28 dimensions, including the state change of each joint (22-dim) and the wrist transformation (6-dim). 
The object's position is randomized for each attempt within a $10cm \times 10cm$ square on the table.

\textbf{\emph{Dexterous Hand Articulated-Manipulation}} aims at grasping and unscrewing the door handle to open the door.
The observation space is 32 dimensions, including the dexterous robot hand state (18-dim), the door handle's position (3-dim) and quaternion (4-dim), and the door base's position (3-dim) and quaternion (4-dim). In contrast, the action space is 28 dimensions. The door's position is randomized for each attempt within a $40cm \times 40cm$ square on the floor.

\textbf{\emph{Dexterous Hand Tool-Use}} aims at picking a hammer and using it to drive a nail into a board.
The observation space is 32 dimensions, including the dexterous robot hand state (18-dim), hammer's position (3-dim) quaternion (4-dim), and nail's position (3-dim). At the same time, the action space is 28 dimensions. The nail's position is randomized for each attempt within a $10cm \times 10cm$ square on the table.

\textbf{\emph{Parallel Gripper Pick-and-Place}} aims at picking an object on the table and placing it into a container.
The observation space is 27 dimensions, including the five rigid bodies of the gripper to object distances (15-dim), the distance between left and right grippers (3-dim), the object's position (3-dim), the distance between object and target (3-dim,) and the distance between flange and target (3-dim).
The action space is 8 dimensions, including the state change of each joint (7-dim) and gripper (1-dim). 
The object's position is randomized for each attempt within a $10cm \times 10cm$ square on the table.

\textbf{\emph{Parallel Gripper Articulated-Manipulation}}
aims at picking an object on the table and placing it into a container.
The observation space is 16 dimensions, including the five rigid bodies of gripper to object distances (15-dim), and the distance between object and target (1-dim).
The action space is 7 dimensions, including the state change of each joint (7-dim). 
The object's position is randomized for each attempt within a $10cm \times 10cm$ square on the table.

\textbf{\emph{Parallel Gripper Cube-Push}}
aims at pushing an object on the table to the target position.
The observation space is 22 dimensions, including the three rigid bodies of the gripper to object distances (9-dim), the flange's position (7-dim), the distance between object and target (3-dim,) and the distance between flange and target (3-dim).
The action space is 7 dimensions, including the state change of each joint (7-dim). 
The object's position is randomized for each attempt within a $5cm \times 5cm$ square on the table.

\subsection{Ablation study}

\begin{figure*}[t]
\setlength{\abovecaptionskip}{0cm}
\setlength{\belowcaptionskip}{0cm}
\centering
\subfigure[\emph{Pick-and-Place}]{
 \begin{minipage}[t]{0.32\linewidth}
    \includegraphics[width=\textwidth]{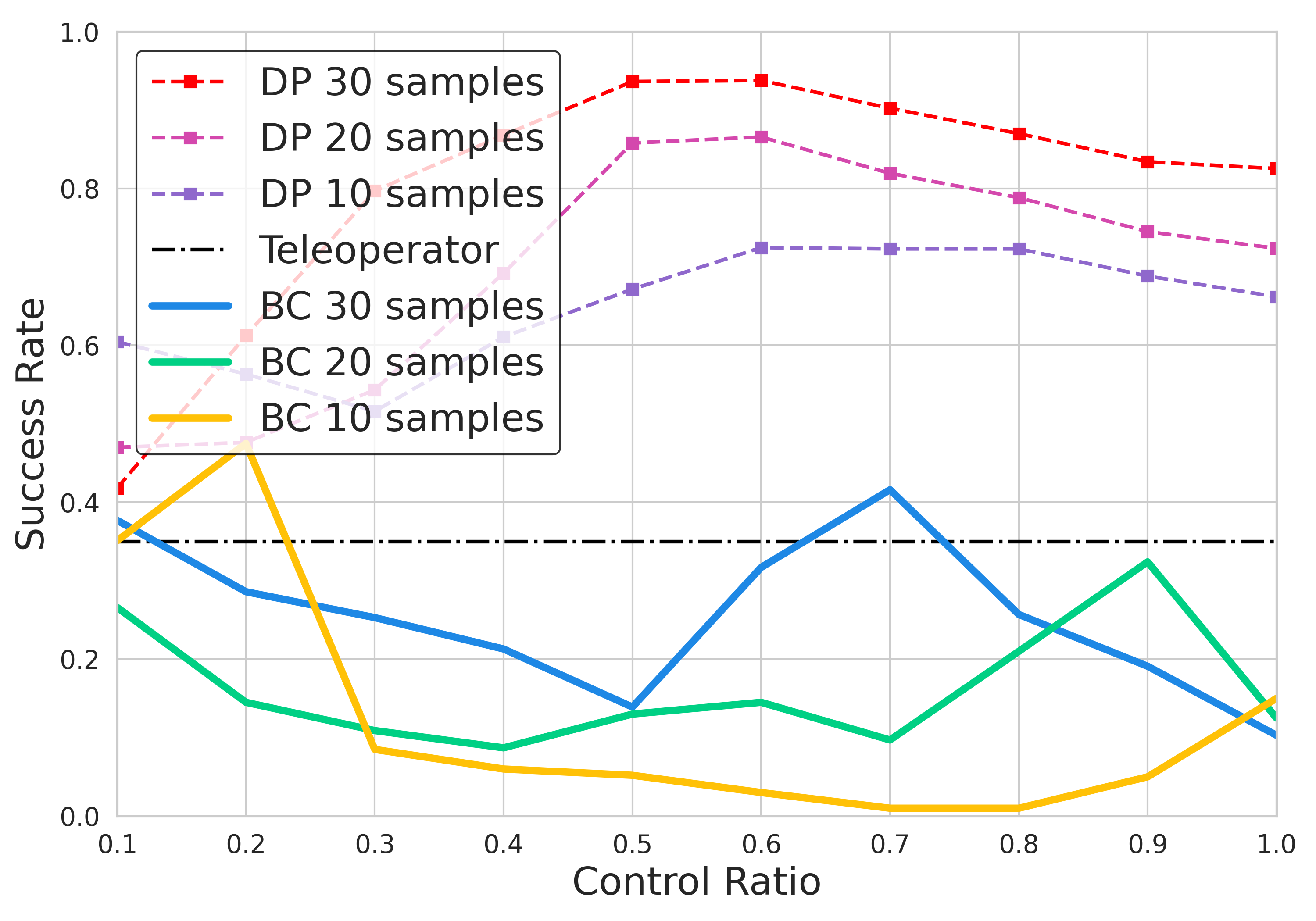}
    \centering
    \end{minipage}}
\subfigure[\emph{Articulated-Manipulation}]{
 \begin{minipage}[t]{0.32\linewidth}
    \includegraphics[width=\textwidth]{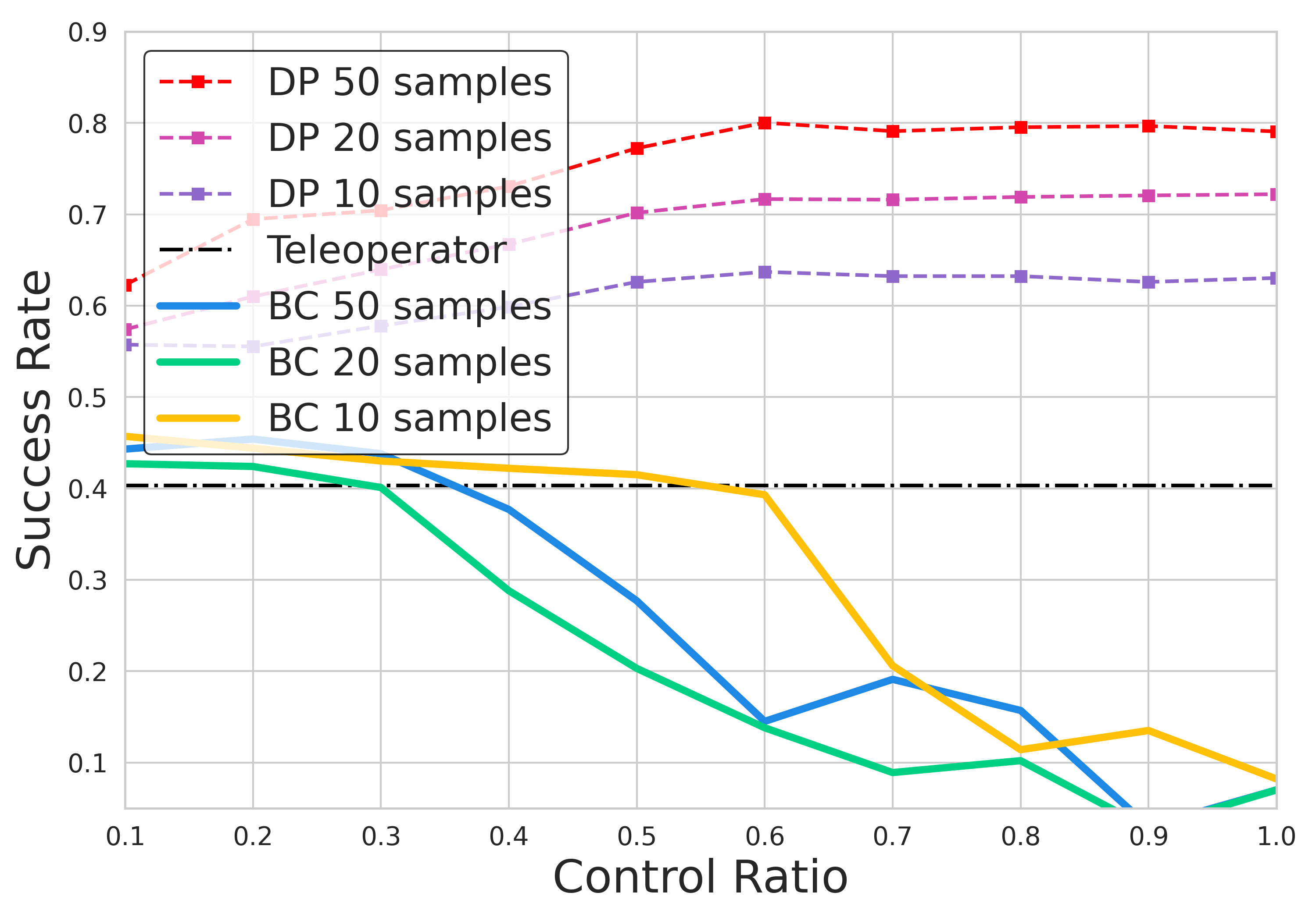}
    \centering
    \end{minipage}}
\subfigure[\emph{Tool-Use}]{
 \begin{minipage}[t]{0.32\linewidth}
    \includegraphics[width=\textwidth]{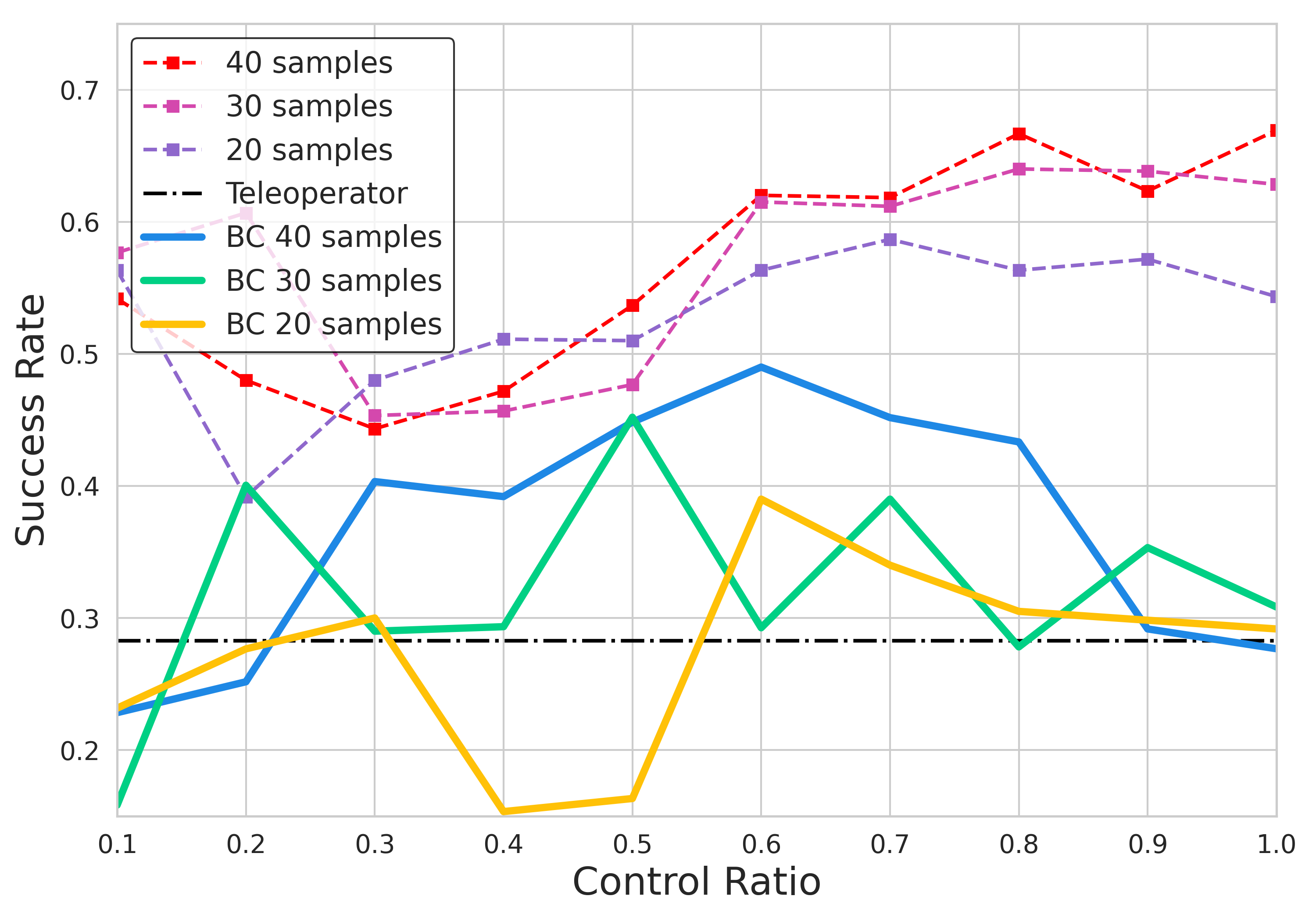}
    \centering
    \end{minipage}}
\caption{Ablation on different dexterous agents trained with different compositions of data.}
\label{fig:ablation}
\end{figure*}

We implement the shared control agent with different methods like the diffusion model and BC. BC adapts a classical way for blending policy to achieve shared control \cite{dragan2013policy}. We use it in the ablation study to blend BC policy with pure human action to achieve shared control in Fig.\ref{fig:ablation}. Compared to the classical way which explicitly averages human action $a^h$ and agent action $a^r$ to get the shared action $a^s$, we instead use the diffusion model, which is a popular implicit model, to blend two actions. It models the process as the forward and reverse process. The forward/diffuse process is about adding Gaussian noise to human action $a^h$, and the reverse process uses a neural network $f(\cdot | \cdot)$ to denoise $a^k$ to get the shared action $a^s$.

BC agent is trained using a specific sequence of data collection and fine-tuning steps to optimize performance across different levels of shared control. Initially, we collect data sets of 10, 10, and 20 episodes under various task conditions. These initial datasets are used to train a preliminary agent. Following this initial training phase, we employ the trained agent to assist in further data collection under three different control ratios represented by $\gamma$ values of 0.25, 0.5, and 0.75. The data collected with the assistance of the agent under these $\gamma$ settings are then used to fine-tune the agent.

As shown in Fig.\ref{fig:ablation}, experiments demonstrated that the success rate of an assistive agent based on BC is lower than that of an agent based on diffusion models, indicating a reduced capacity for assistance. In certain instances, the action even becomes worse at particular control ratios. 

\begin{table}[t!]
\setlength{\abovecaptionskip}{0cm}
\setlength{\belowcaptionskip}{0cm}
\caption{Agent performance on human expert or amateur datasets.}
\begin{center}
    \resizebox{0.48\textwidth}{!}{
\begin{tabular}{c|cc|cc|cc}
\toprule
Dexterous& \multicolumn{2}{c|}{\emph{Pick-and-Place}} & \multicolumn{2}{c|}{\emph{Articulated-Manipulation}} & \multicolumn{2}{c}{\emph{Tool-Use}}\\
Hand& \emph{Skilled}& \emph{Unskilled}& \emph{Skilled}& \emph{Unskilled}& \emph{Skilled}& \emph{Unskilled}\\
\midrule\midrule
\emph{BC} & 0.45& 0.02& 0.43& 0.18& 0.40& 0.05\\
\emph{BC-RNN} & 0.41& 0.05& 0.62& 0.04& 0.27& 0.05\\
\emph{DP} & 0.71& 0.01& 0.68& 0.10& 0.81& 0.03\\
\bottomrule
\end{tabular}}
\end{center}
\label{tab:humanablation}
\end{table}

\begin{table}[t]
\setlength{\abovecaptionskip}{0cm}
\setlength{\belowcaptionskip}{0cm}
\caption{Ablation study on DP performance between $r$.}
\begin{center}
\begin{tabular}{c|ccc}
\toprule
& \emph{Pick-and-Place}& \emph{Articulated-Manipulation}& \emph{Tool-Use}\\
\midrule\midrule
$r = 0.0$& 0.565& 0.661& 0.512\\
$r = 0.1$& \textbf{0.620}& \textbf{0.681}&\textbf{0.547}\\
$r = 0.2$& 0.575& 0.407& 0.115\\
$r = 0.3$& 0.435& 0.216& 0.029\\
\bottomrule
\end{tabular}
\end{center}
\label{tab:noiseablation}
\vspace{-20px}
\end{table}

We test the performance of training with different data compositions.
For a task, we gathered two manipulation datasets from both skilled and unskilled human operators. We consider operators to be skilled workers if they can practice for more than five hours and reach a success rate and efficiency comparable to those with assistive agents. As shown in Tab.~\ref{tab:humanablation}, the performance of agents trained on the dataset of unskilled operators is much lower than that on the dataset of skilled operators. Therefore, all the human operation datasets $\mathcal H$ we use in the main text are from skilled operators.

In our framework, $r$ represents the modification ratio of noise between the state and action. Specifically, during the training, the noise added to state $s$ satisfies $\epsilon_s=u\cdot \mathcal{N}(\mathbf{0, I})$ while the noise added to action $a$ satisfies $\epsilon_s=v\cdot \mathcal{N}(\mathbf{0, I})$. Then $r=\frac u v$. We test different $r$ as shown in Tab.~\ref{tab:noiseablation}, to ensure the best agent performance. We default to using $r = 0.1$ in our model.

\begin{figure}[t!]
    \centering
    \includegraphics[width=1\linewidth]{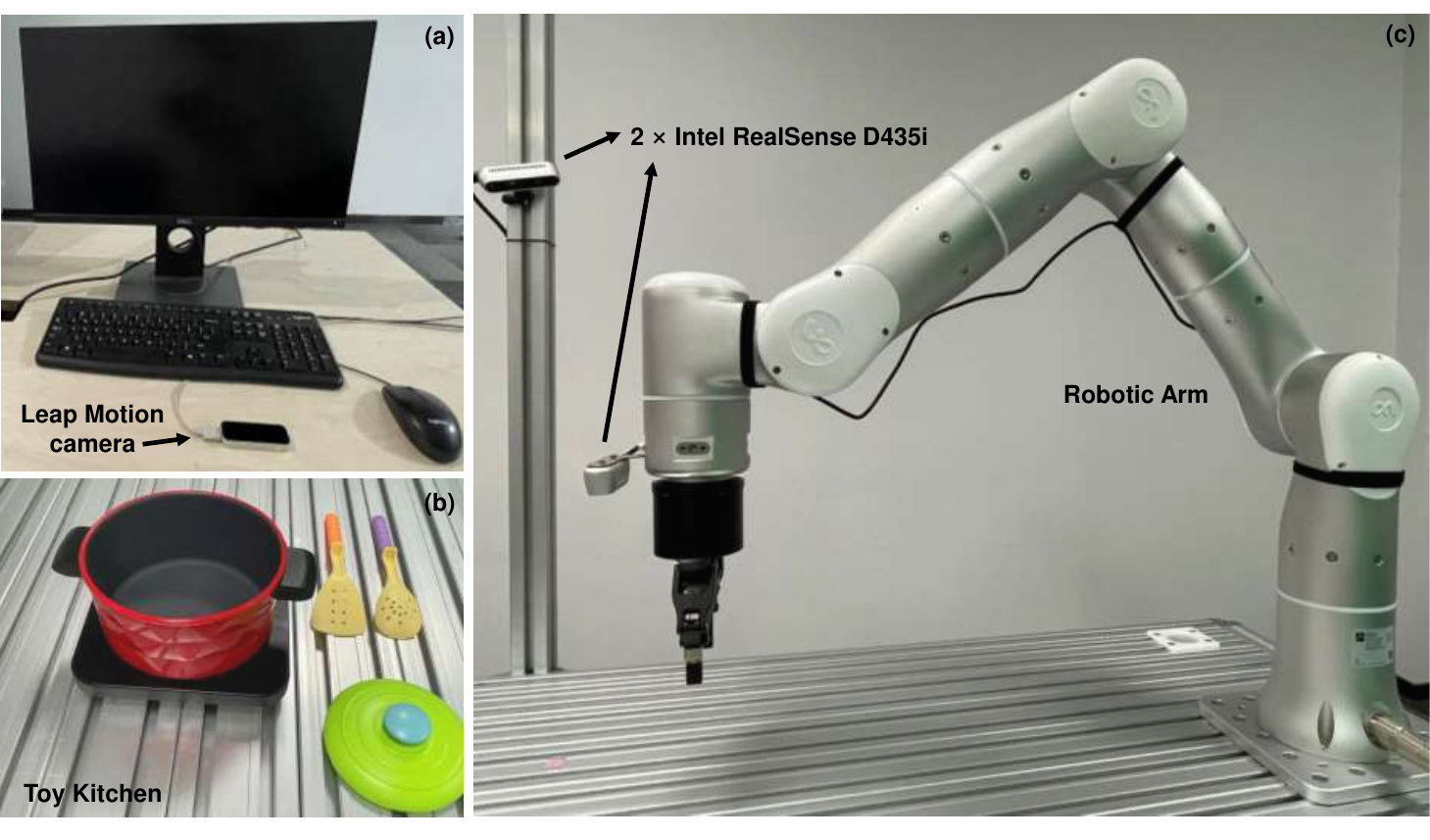}
    \caption{Realworld Pick-and-Place Experiment. The hardware setup comprises (a) a Leap Motion camera utilized for teleoperation data collection, (b) a toy kitchen environment set up for the pick-and-place task, and (c) a Flexiv Rizon4 robotic arm equipped with a gripper and two cameras. One camera is mounted on the wrist of the robotic arm, while the second one is positioned on the side.}
    \label{fig:real_world_scene}
\end{figure}

\subsection{Real World Experiment}

In this section, we evaluate the real-world performance of our method. We use the setup shown in Fig.\ref{fig:real_world_scene}, which includes a Flexiv Rizon4 arm equipped with a gripper and two Intel RealSense D435i RGB-D cameras. One camera is mounted on the wrist of the robotic arm, while the second is positioned on the side. One task here is to pick the red pot shown in Fig.\ref{fig:real_world_scene} and place it onto the induction cooker.


During the real-world data collection phase, we estimate the human hand's pose using RGBD input. Considering the significant difference in morphology between the human hand and a 7-DoF robotic arm, we chose to track the end effector's position by monitoring the position of the hand's wrist. Additionally, we used the action of closing or opening the human hand as the condition for determining whether to grasp or release an object. This approach leverages the greater dexterity of the human hand to enhance the control and precision of the robotic arm.  We record RGB images from two camera views, joint poses (7-dim), gripper width (1-dim), the end effector's position (3-dim), and its quaternion (4-dim). The RGB images have a size of $640\times 480$ pixels, each episode is sampled at a frequency of 10 Hz.

In real-world experiments, the network architecture is generally similar to the simulation environment's. Our input has changed from the original hand states and object states to the position and orientation of the robot arm end effector, as well as images from the first-person and third-person perspectives. We made two main modifications:
1) For the images, we used a ResNet-18 model. We used a standard ResNet-18 (without pretraining) as the encoder with its global average pooling replaced with a spatial softmax pooling to maintain spatial information.
2) We deepened the layer of the neural network, increased its hidden layer dimension, and
 expanded action horizon prediction from predicting the next frame action to predicting actions for the subsequent $T$ frames, \textit{i.e.}, $a_{t+1:t+T-1}$ (where $T=8$).

\section{Discussion and Limitation}
\subsection{Human-Machine Interface}
Our approach has demonstrated success across a diverse set of Human Machine Interfaces(HMI), including:

\textbf{Sigma.7 Teleoperation Devices}: Our system has successfully utilized Sigma devices to achieve precise control for tasks involving limited DoF. These devices require intricate control and feedback mechanisms, demonstrating our interface's robustness and effectiveness in physical UI scenarios.

\textbf{RGB-D Cameras}: Our system can accurately interpret spatial environments by leveraging depth perception, making it highly effective for freehand teleoperation. This capability lays the foundation for handling physical UIs with equal precision.

\textbf{Virtual Reality (Meta Quest3)}: In VR environments, our interface provides an immersive and intuitive experience that closely mimics real-world interactions. This shows its capability to handle complex interfaces with precision and ease. As shown in Tab.~\ref{tab:VR}, we repeated the dexterous articulated-manipulation experiment with Leap Hand~\cite{shaw2023leap} in a VR environment and validated that our paradigm is applicable across different HMIs. This demonstrates the versatility of our approach, ensuring consistent operation across various human-machine interfaces.

\begin{table}[h!]
\setlength{\abovecaptionskip}{0cm}
\setlength{\belowcaptionskip}{0cm}
    \centering
\captionof{table}{Articulated-Manipulation task success rate under increasing data with Quest3.}
\begin{tabular}{c|cc}
    \toprule
VR    & \multicolumn{2}{c}{\emph{Articulated-Manipulation}}\\
Dexterous    &\emph{BC} & \emph{DP} \\
    \midrule
    10$\mathcal H$ & 0.04 & 0.10\\
    \midrule
    10$\mathcal H$ + 10$\mathcal H$& 0.15 & 0.25\\
    10$\mathcal H$ + 20$\mathcal H$& 0.26 & 0.26 \\
    10$\mathcal H$ + 30$\mathcal H$& 0.40 & 0.30 \\
    \midrule
    10$\mathcal H$ + 10$\mathcal S$ & 0.34 & 0.28 \\
    10$\mathcal H$ + 20$\mathcal S$ & 0.30 & 0.35 \\
    10$\mathcal H$ + 30$\mathcal S$ & 0.44 & 0.63 \\
    \bottomrule
\end{tabular}
\label{tab:VR}
\end{table}

\begin{figure*}[t!]
\setlength{\abovecaptionskip}{0cm}
\setlength{\belowcaptionskip}{0cm}
\centering
\subfigure[\emph{Pick-and-Place}]{
 \begin{minipage}[t]{0.32\linewidth}
    \includegraphics[width=\textwidth]{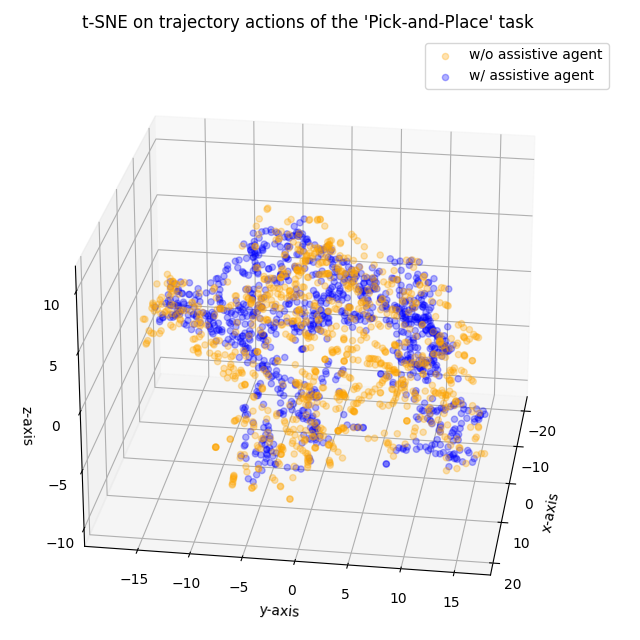}
    \centering
    \end{minipage}}
\subfigure[\emph{Articulated-Manipulation}]{
 \begin{minipage}[t]{0.32\linewidth}
    \includegraphics[width=\textwidth]{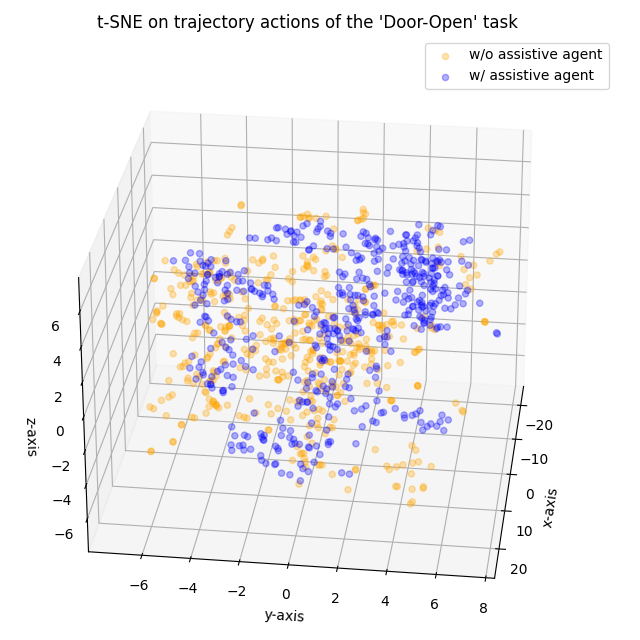}
    \centering
    \end{minipage}}
\subfigure[\emph{Tool-Use}]{
 \begin{minipage}[t]{0.32\linewidth}
    \includegraphics[width=\textwidth]{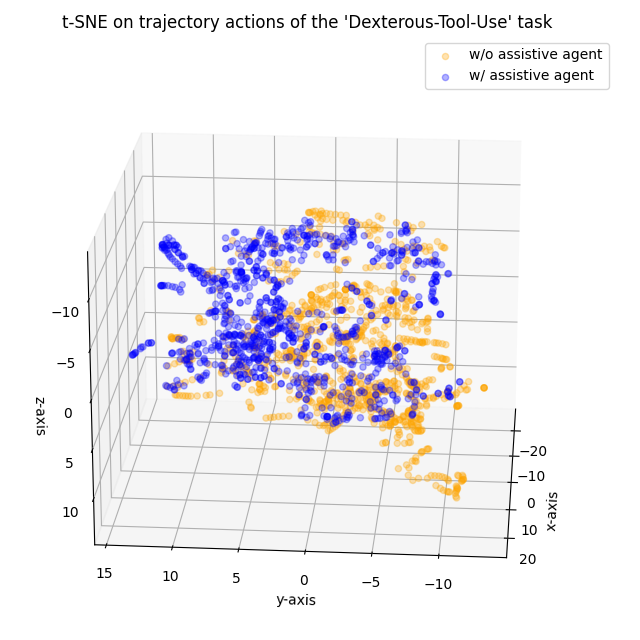}
    \centering
    \end{minipage}}
\caption{t-SNE visualization of distributions of trajectory actions (w/ and w/o assistive agent).}
\label{fig:tsne}
\vspace{-10px}
\end{figure*}

\subsection{Data Analysis}
\begin{figure}[h!]
    \centering
    \includegraphics[width=0.8\linewidth]{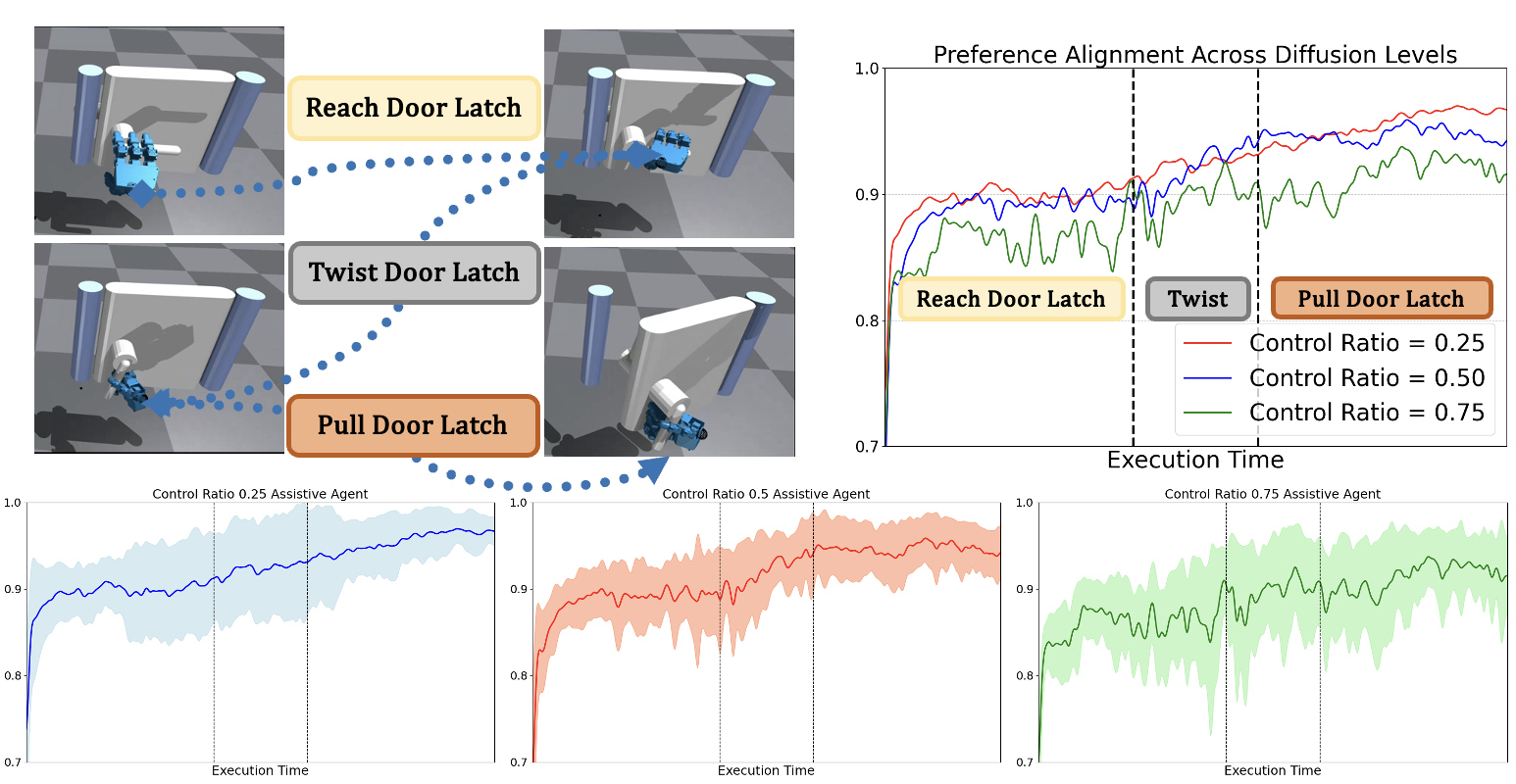}
    \caption{The articulated-manipulation task consists of three phases: reaching the door latch, twisting it to the correct angle, and pulling it. We plotted the dot product between the user input action and the assistive agent's output action(Preference Alignment). In the plot, the red, green, and blue lines represent control ratios of 0.25, 0.5, and 0.75, respectively.}
    \label{fig:analysis}
\end{figure}

We visualize the Preference Alignment~\cite{jeon2020shared} for dexterous hand articulated-manipulation task, as shown in Fig.~\ref{fig:analysis}. We find that as time progresses, the preference alignment increases across all three phases: reaching the door latch, twisting the door latch, and pulling. This indicates a growing synchronization between the user and the assistive agent throughout each stage of the task. Also, the preference alignment between the user and the assistive agent improves across different control ratios. 

We use t-SNE to visualize the distribution of trajectory actions on different dexterous tasks, as shown in Fig.~\ref{fig:tsne}. Specifically, we have reduced the trajectory of actions to three dimensions using t-SNE, for both data collected by human operators with and without our system. To ensure a fair comparison, we uniformly sampled the same number of actions across both scenarios. We find that the distribution of the same task tends to cluster in the same space, whether with or without an assistive agent. This indirectly demonstrates that our system can enhance data collection speed and efficiency without compromising data quality.

\subsection{Limitations}
Our current system's task-specific assistive agent, while effective for certain applications, does have its limitations. It currently can not handle tasks that involve multiple subtasks or targets that change dynamically, as these scenarios often require more flexibility, including the ability to adjust the control ratio throughout the sequence. To broaden the system's applicability, integrating large language models could also allow it to handle a wider range of robot learning tasks by conditioning on text input. Additionally, we think adding a learnable control ratio adjustment mechanism, especially for long-horizon tasks, could improve the system's adaptability and efficiency. We believe that our proposed joint-learning framework has the potential to leverage more powerful multi-task diffusion policies, allowing it to handle more complex scenarios in future enhancements.

\section*{ACKNOWLEDGMENT}

The authors would like to thank anonymous reviewers for helping improve exposition. This work is supported in part by the National Natural Science Foundation of China under Grants 62306175.



\bibliographystyle{IEEEtranN}
\bibliography{references}